\pdfoutput=1

\documentclass[11pt]{article}

\usepackage[final]{acl}

\usepackage{times}
\usepackage{latexsym}

\usepackage[T1]{fontenc}

\usepackage[utf8]{inputenc}

\usepackage{microtype}

\usepackage{inconsolata}

\usepackage{graphicx}

\usepackage{graphicx}
\graphicspath{ {./images/} }
\usepackage{import}

\usepackage{adjustbox}
\usepackage{array}
\usepackage{multirow}
\usepackage{tikz}
\usepackage{siunitx}
\usepackage{booktabs}
\usepackage{caption}
\usepackage{acronym}
\usepackage{tabularx}
\usepackage{geometry}
\usepackage{svg}
\usepackage{xcolor}  
\usepackage{color}   
\usepackage{colortbl} 
\usepackage{float}
\usepackage{orcidlink}
\usepackage{longtable}
\usepackage{ragged2e}
\usepackage{tablefootnote}
\usetikzlibrary{shapes.geometric, arrows, shadows}

\tikzstyle{startstop} = [rectangle, rounded corners, minimum width=2cm, minimum height=0.8cm, text centered, draw=black, fill=red!30, drop shadow]
\tikzstyle{process} = [rectangle, minimum width=2cm, minimum height=0.8cm, text centered, draw=black, fill=blue!20, drop shadow]
\tikzstyle{decision} = [diamond, minimum width=2.5cm, minimum height=0.8cm, text centered, draw=black, fill=green!30, drop shadow]
\tikzstyle{arrow} = [thick,->,>=stealth, draw=black!80]

\definecolor{derailmentcolor}{HTML}{CCE5FF}
\definecolor{tacticscolor}{HTML}{DFF2BF}
\definecolor{othercolor}{HTML}{FFECB3}
\definecolor{facilitationcolor}{HTML}{D3D3F9}
\definecolor{taxonomycolor}{HTML}{E6F2D6}

\definecolor{lightblue}{RGB}{220,230,241}
\definecolor{peach}{RGB}{255,218,185} 
\definecolor{mintgreen}{RGB}{199, 233, 192} 
\definecolor{blushpink}{RGB}{255, 205, 210}
\definecolor{pastelblue}{RGB}{204, 229, 255}

\newcommand{\taskderailment}{\cellcolor{derailmentcolor} Conversation Derailment}

\newcommand{\taskfacilitation}{\cellcolor{facilitationcolor} Facilitator Interventions}

\newcommand{\KKorcid}{\orcidlink{0000-0002-9349-9554}}
\newcommand{\Dimorcid}{\orcidlink{0000-0002-5675-3939}}
\newcommand{\NGorcid}{\orcidlink{0000-0002-7797-2030}}
\newcommand{\DMorcid}{\orcidlink{0009-0007-3701-1458}}

\title{Evaluation and Facilitation of Online Discussions in the LLM Era: \\A Survey}

\author{
    Katerina Korre$^{\dagger}$\KKorcid, Dimitris Tsirmpas$^{\dagger\ddagger}$\Dimorcid, Nikos Gkoumas$^{\dagger}$\NGorcid, Emma Cabalé$^{\diamondsuit}$\orcidlink{ 0009-0009-2230-9408}\\
    \textbf{Danai Myrtzani}$^{\ddagger}$\DMorcid, \textbf{Theodoros Evgeniou}$^{\spadesuit}$\\
    \textbf{Ion Androutsopoulos}$^{\ddagger\dagger}$, \textbf{John Pavlopoulos}$^{\ddagger \dagger}$\orcidlink{0000-0001-9188-7425} \\
    \small
    $^{\dagger}${Archimedes, Athena Research Center, Greece} (\texttt{\{k.korre,n.goumas\}@athenarc.gr})\\
    \small
    $^{\ddagger}$Athens University of Economics and Business, Greece (\texttt{\{dim.tsirmpas,dan.myrtzani,ion,annis\}@aueb.gr})\\
    \small
    $^{\diamondsuit}$École Normale Supérieure Paris-Saclay, France (\texttt{emma.cabale@ens-paris-saclay.fr})\\
    \small
    $^{\spadesuit}$INSEAD, Technology and Business, France (\texttt{theodoros.evgeniou@insead.edu})\\
  }

\begin{document}
\maketitle
\begin{abstract}
We present a survey of methods for assessing and enhancing the quality of online discussions, focusing on the potential of \acfp{LLM}. While online discourses aim, at least in theory, to foster mutual understanding, they often devolve into harmful exchanges, such as hate speech, threatening social cohesion and democratic values. Recent advancements in \acp{LLM} enable artificial facilitation agents to not only moderate content, but also actively improve the quality of interactions. 
Our survey synthesizes ideas from \acf{NLP} and Social Sciences to provide (a) a new taxonomy on discussion quality evaluation, (b) an overview of intervention and facilitation strategies, (c) along with a new taxonomy of conversation facilitation datasets, (d) an \ac{LLM}-oriented roadmap of good practices and future research directions, from technological and societal perspectives.
\end{abstract}

\section{Introduction}


Discussions, especially of complex or controversial topics, are a cornerstone of collective decision-making \cite{burton2024large}. 
In contrast to initial hopes of promoting mutual understanding 
\cite{rheingold00},
online discussions (especially in social media) often degenerate into hate speech, personal attacks, promoting conspiracy theories or propaganda -- to the extent that they can even be considered a threat to social cohesion and democracy \cite{tucker2018social,binny2019}.

Natural Language Processing (\ac{NLP}) and \ac{ML} can potentially help improve the quality of online discussions.
For example, 
automatic classifiers 
\cite{bang-etal-2023-enabling,Molina2022} 
are already being used to help or even replace human moderators, by flagging 
posts that violate the law or 
policies of online discussion fora \cite{saeidi-etal-2021-cross}. 

Social Science provides theories and applications for the facilitation of a discussion, but in specific contexts, such as teaching \cite{Mansour2024} or clinical discussions \cite{Gelula1997},
without much research devoted to online discussions. While prior \ac{NLP} studies have explored \ac{LLM}-facilitated discussions \cite{burton2024large, Aher2023, beck-etal-2024-sensitivity, schroeder-etal-2024-fora, small-polis-llm, cho-etal-2024-language}, rarely does Social Science work examine how facilitation can be automated \cite{Gimpel2024}.







\begin{figure}[!t]
    \centering
    \includegraphics[width=\linewidth]{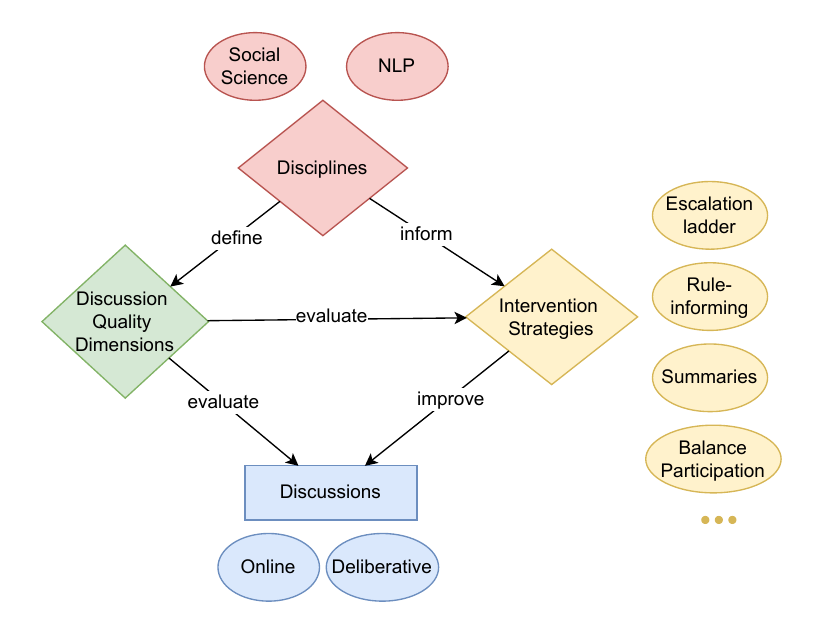}
    \caption{
    A conceptualization of this survey. We explore approaches from different disciplines, which recommend their own ways of evaluating and improving discussions.}
    \label{fig:survey}
\end{figure}

In this survey, we combine \ac{LLM}-based methods, with ideas from Social Science (e.g., 
Deliberative Theory)
when discussing 
how to evaluate online discussions, and when exploring intervention strategies. 
Figure~\ref{fig:survey} 
provides a high-level conceptualization of our work.

The main research question of this survey is \emph{can \acp{LLM} be used effectively as facilitators in online discussions?} Focusing on threaded discussions (\S\ref{sec:terminology}), we explore three key areas: (1) methods (potentially also \ac{LLM}-based) for evaluating aspects of online discussions, (2) intervention strategies for facilitation, and (3) available data resources which can be used to analyze human facilitation and train \ac{LLM} equivalents.
Specifically,
we survey discussion evaluation aspects and introduce a new taxonomy (\S\ref{sec:eval_methods}).
We map tasks suited for \ac{ML} models, \acp{LLM}, and humans, aggregate multidimensional insights on facilitation strategies (\S\ref{sec:intervention_strategies}), and outline future possibilities for \acp{LLM} (\S\ref{sec:towards_llm_facilitation}). Additionally, we aggregate and compare all major relevant datasets in literature, dividing them into categories per task (\S\ref{sec:datasets}).

Our findings show that (a) many discussion evaluation dimensions coexist in the literature; 
(b) \ac{LLM} advancements show significant promise in improving the quality and timeliness of facilitation methods; (c) while surveying the existing datasets, we notice a scarcity of datasets for studying facilitation. We posit that \ac{LLM}-generated discussions, could become an asset to develop and test automatic facilitation strategies in diverse artificial discussions, before testing the strategies and the \ac{LLM}-based facilitator agents 
in more costly experiments with human participants. 



\section{Terminology}
\label{sec:terminology}

Given the numerous aspects to consider regarding discussion quality and facilitation, we clarify the terminology we use. We highly recommend consulting the Terminology Section of Appendix~\ref{sec:appendix:terminology} and, especially,  Table~\ref{tab:terminology}, where we explain our findings with regard to the terms used in the literature.

\paragraph{Facilitation vs.\ Moderation}
The term `moderation' is more commonly used in \ac{NLP} \cite{argyle2023}, typically referring to the flagging and/or removal of unwanted content (`content moderation'), while `facilitation' is more prevalent in the Social Sciences, where it encompasses a broader scope, including active interventions \cite{vecchi-etal-2021-towards, kaner2007facilitators, trenel2009facilitation}. Given the limited attention to facilitation in \ac{NLP} and the survey’s grounding in Social Science, we distinguish between the terms, even though they are sometimes used interchangeably in the literature.


\paragraph{Ex-Post moderation} This survey mainly focuses on `Real-Time, Ex-Post-moderation', i.e., moderation happening just after the user has posted some content. This is different from pre-moderation approaches, 
such as nudging users before they post harmful content \cite{argyle2023}, or delaying the posting of user content until a moderator has had the chance to check it.

\paragraph{Discussion, Deliberation, Dialogue, Debate} 
The definitions of these terms often vary across literature \cite{Russman16, Goni2024}. We focus on \textbf{discussions}, a general term for verbal/written exchanges \cite{Russman16},  and \textbf{deliberations}, a term for structured discussions focusing on opinion sharing \cite{DEGELING2015114, Lo_McAvoy_2023}. This is in contrast to the (at least in theory) collaborative nature of \textbf{dialogues} \cite{Redwood18, bawden-2021-understanding, Goni2024} and the competitive and organized nature of \textbf{debates} \cite{Lo_McAvoy_2023}.

\paragraph{Tree-style discussions} (or `threads') are discussions which start from an Original Post (OP) with subsequent comments replying to either the OP or to other comments \cite{seering_self_moderation}.

\section{Comparison to Other Surveys}
Only two studies have surveyed the field of \ac{NLP} while also considering ideas from Social Science. However, they focus mainly on \ac{AM}. These are the studies of \citet{wachsmuth-etal-2024-argument} and \citet{vecchi-etal-2021-towards}. \citet{wachsmuth-etal-2024-argument} focus primarily on discussion \emph{evaluation} disregarding its relation to facilitation, which is one of the main goals of our survey. 
The survey of \citet{vecchi-etal-2021-towards} argues that advancing \ac{AM} for social good requires a collaborative effort between \ac{AM} and Social Science.
They point out that traditional \ac{AM} has prioritized the logical structure and soundness of arguments, while overlooking other important dimensions, such as civility, respectfulness, inclusiveness, originality, and the broader impacts of discussions—such as encouraging mutual understanding and problem-solving. Building on these notions, we incorporate ideas from Social Science into \ac{NLP}-based approaches, discussing both discussion evaluation and facilitation, both with a focus on the potential of \acp{LLM}.

\section{Discussion Quality Evaluation}
\label{sec:eval_methods}
Improving online discussions presupposes being able to define and measure \emph{discussion quality}. 
While there have been attempts to provide frameworks for discussion quality evaluation \cite{kie2022online,gerber2018deliberative}, none of them is directed towards facilitation. 
Crucially, most existing frameworks ultimately rely on human judgments as their reference point, yet human evaluation is expensive, slow, and shows low inter-rater agreement on dimensions that involve subjective interpretation, such as pragmatic cues \citep{smith-etal-2022-human, yeh-etal-2021-comprehensive, khalid-lee-2022-explaining}. This evaluation bottleneck motivates a taxonomy of evaluation methods that is both comprehensive and amenable to scalable automatic measurement.


In this work, we draw from the works of \citet{10.1093/oso/9780192848925.003.0006, https://doi.org/10.1111/j.1467-9760.2009.00342.x,steenbergen2003measuring,falk2023bridging} and \citet{kie2022online} 
to define a new social-science-informed taxonomy for discussion quality dimensions. While we present a structured taxonomy, it is important to note that the categories are not mutually exclusive. Rather, elements within the taxonomy may coexist within evaluation dimensions, complement one another, or serve as explanatory mechanisms for other dimensions. An example of the dimension interaction can be found in Table~\ref{app:tab:eval_example} in the Appendix~\ref{app:eval_example}. The grouped dimensions along with the \ac{NLP} approaches are shown in the Appendix in Table~\ref{tab:discussion_quality_grouped_alt}.


\subsection{Structure and Logic}\label{subsec:structure_logic}

\paragraph{Argument Structure and Analysis} \ac{AQ} is a multidimensional concept assessed through logical, rhetorical, and dialectical dimensions \cite{wachsmuth2017computational}. The logical dimension focuses on the coherence and structure of the argument. The rhetorical dimension assesses persuasiveness, focusing on the argument's style and emotional appeal. The dialectical dimension assesses the constructiveness of the argument. Empirically, threads with well-formed claim-evidence chains exhibit higher coherence and lower odds of devolving into ad-hominem attacks, making \ac{AQ} scores, as a discussion quality dimension, an \emph{early-warning} indicator of derailment \citep{chang-danescu-niculescu-mizil-2019-trouble}.
All the above dimensions of automatic argument-structure analysis can be used
by a facilitator to keep the discussion fact-centered, inclusive, and on track \cite{falk-etal-2021-predicting, falk2023bridging}.



\paragraph{Coherence and Flow} Coherence, as described above, evaluates logical consistency, while flow assesses smooth progression in discussions \cite{li-etal-2021-conversations}. Both are essential tools for facilitators in their effort to redirect off-topic comments and guide transitions between topics during a discussion \cite{Lambert2024,park_et_al_2012_facilitation,falk-etal-2024-moderation}. A sudden drop in how well responses match the topic or question often comes before personal attacks or off-topic turns \citep{chang-danescu-niculescu-mizil-2019-trouble, zhang-etal-2018-conversations}, making coherence and flow indicators of argument structure and a valuable early signal for facilitators.



\paragraph{Turn-taking} How speakers alternate, the frequency of their turns, and the participants they address can serve as a diagnostic of conversational health. Balanced exchanges enhance coherence \citep{cervone-riccardi-2020-dialogue}, predict constructiveness (\S\ref{subsec:emotion_behavior}) \citep{niculae-danescu-niculescu-mizil-2016-conversational}, and provide facilitators with actionable cues \citep{schroeder-etal-2024-fora}. To gauge speaking time, turn count, and word usage, researchers have applied metrics such as entropy \cite{niculae-danescu-niculescu-mizil-2016-conversational} and Gini coefficients \cite{schroeder-etal-2024-fora}.


\paragraph{Linguistic Markers} Linguistic markers 
have been used to help model content and expression in online discussions  \cite{doi:10.1177/0261927X8400300301}. Early methods used lexicons for sentiment, toxicity, politeness (\S\ref{subsec:social_dynamics} and~\ref{subsec:emotion_behavior}) and collaboration evaluation
\cite{10.1145/3032989, avalle2024persistent}. 
For example, spikes in hedges (e.g., `maybe', `I guess') invite clarification requests by facilitators, while bursts of second-person pronouns, similarly to turn-taking, often foreshadow personal attacks and can prompt a civility nudge \cite{niculae-danescu-niculescu-mizil-2016-conversational}.

\paragraph{Speech and Dialogue Acts} 
Rooted in Speech Act Theory \cite{austin1975things, Searle_1969}, dialogue acts have been
employed to assess deliberative quality and analyze facilitation strategies \cite{Fournier-Tombs21,chen2024exploring}. They characterize dialogue turns (e.g., interruption) to analyze interaction dynamics \cite{ferschke-etal-2012-behind, stolcke-etal-2000-dialogue, Zhang_Culbertson_Paritosh_2017, al-khatib-etal-2018-modeling}. 
Positive (e.g., causal reasoning) or negative (e.g., disrespect) dialogue acts can be scored to reflect discussion quality with low scores potentially indicating a need for intervention \cite{ziems2024can, cimino-etal-2024-coherence, martinenghi-etal-2024-llms, schroeder-etal-2024-fora}.



\paragraph{Pragmatic Comprehension}
Pragmatic comprehension—how context shapes meaning—is crucial to facilitation, as intended meanings often diverge from literal expressions (i.e., implicature). Humans resolve such ambiguity using social and commonsense knowledge. Grice’s maxims \cite{grice1975logic}, a central pragmatic concept, can help explain this process by outlining the conversational principles people rely on to infer meaning, while they have already been used to assess discussion quality \cite{jwalapuram-2017-evaluating,10.1145/3411764.3445312,NGAI2021101098,doi:10.1089/cyber.2022.0356}.

\subsection{Social Dynamics}\label{subsec:social_dynamics}

\paragraph{Politeness}

Politeness serves as a cornerstone of prosocial behavior, an attribute that facilitators desire to foster in online discussion forums \cite{Lambert2024}. In the context of facilitation, it has mainly been studied in relation to conversational derailment (\S\ref{sec:datasets}) \cite{zhang-etal-2018-conversations} and constructiveness (\S\ref{subsec:emotion_behavior}) \cite{de-kock-vlachos-2021-beg,zhou-etal-2024-llm-feature}. 




\paragraph{Power and Status}

Power and status influence conversational dynamics, affecting language use and turn-taking (\S\ref{subsec:structure_logic}). Higher status speakers can control the flow of discussions and foster social inequalities. Interestingly, low-status individuals tend to mimic 
the linguistic styles of high-status speakers more than 
the opposite
\cite{10.1145/2187836.2187931}, and this can be used as a signal that there is high/low-status imparity in a discussion. 
Facilitators may intervene, then, to ensure that the right to speak is evenly distributed among participants, preventing projection of social biases and stereotypes. 



\paragraph{Disagreement}

Disagreements, when constructive, improve discussions by fostering deeper understanding
\cite{friess2018letting,de-kock-vlachos-2021-beg}. Assessing disagreement, however, is complex. The hierarchy of \citet{graham2008disagree} considers disagreement tactics ranging from name calling to refuting the central point. Along with other work on dispute tactics 
\cite{walker-etal-2012-corpus,benesch2016counterspeech,de-kock-vlachos-2022-disagree}, it can be used to examine types of disagreements in a discussion.

\subsection{Emotion and Behavior}\label{subsec:emotion_behavior}

\paragraph{Empathy}

Empathy is the ability to understand other perspectives and emotions and respond correspondingly \cite{Lipman_2003,xu-jiang-2024-multi}. Facilitators desire to foster empathy in online discussions, since it encourages prosocial behavior and boosts engagement \cite{xu-jiang-2024-multi,Concannon2024,Lambert2024}. To do so, they encourage users to share personal stories and experiences \cite{schroeder-etal-2024-fora}. Various coding schemes \cite{MACAGNO2022116}, psychological indicators (e.g., the emotion-laden words of \citealp{furniturewala-jaidka-2024-empaths}), and dimensions (e.g., perceived engagement such as in \citealp{xu-jiang-2024-multi}) have been used to detect both expressed and perceived empathetic traits. 

\paragraph{Toxicity} Toxicity in online discussions refers to harmful or disrespectful language that hinders productive discourse and can derail meaningful discussions \cite{avalle2024persistent}. Facilitation is key to maintaining healthy communication, requiring both early detection of toxicity and (in the case of more active facilitation) proactive de-escalation strategies, such as conversation redirection or positive engagement (\S\ref{sec:intervention_strategies}). In the case of conventional moderation that only aims to flag or remove toxic content, debate persists over what content warrants removal \cite{WARNER2025103468, habibi2024contentmoderatorsdilemmaremoval,PRADEL_ZILINSKY_KOSMIDIS_THEOCHARIS_2024}.




\paragraph{Sentiment} Sentiment analysis helps identify whether discussions are positive, negative, or neutral. In the context of facilitation, sentiment analysis gauges the tone of discussions, which influences the quality of interactions \cite{de-kock-vlachos-2021-beg}. Positive sentiment contributions in online discussion forums usually signal prosocial behavior and hence are highly encouraged by facilitators \cite{Lambert2024}, while negative sentiments among discussants contribute to conversation toxicity \cite{avalle2024persistent}. 


\paragraph{Controversy} 
Controversy arises from divergent viewpoints, leading to polarized exchanges that can escalate to toxicity and derail online discussions \cite{avalle2024persistent}. Controversial comments have been shown to contribute to a decline in positive emotions and a sustained rise in anger \cite{hessel-lee-2019-somethings,chen-etal-2025-whow}. The spread of political leanings among discussants and sentiment distribution analysis are common approaches to measure controversy \cite{avalle2024persistent}. 





\paragraph{Constructiveness} Constructiveness fosters meaningful dialogue, especially in online discussions, by promoting resolution and cooperation \cite{shahid2024examining}. 
It is often
signalled by linguistic markers (\S\ref{subsec:structure_logic}) \cite{de-kock-vlachos-2022-disagree, falk-etal-2024-moderation}.
A facilitator can exploit a constructiveness score;  
threads trending upward are worth highlighting or summarizing, whereas a downward drift may trigger facilitation tactics such as slower, structured turn-taking or clarification prompts \citep{de-kock-vlachos-2021-beg}.

\subsection{Engagement and Impact}\label{subsec:engagement_impact}

\paragraph{Engagement} Engagement is 
desirable in online discussion platforms as it combines 
interest and participation \cite{Lambert2024,park_et_al_2012_facilitation}. It is proxied by measures like reciprocity \cite{graham2003,Stromer-Galley2007,zhang-etal-2018-conversations}, number of comments posted by each user \cite{avalle2024persistent}, discussion length \cite{adomavicius2021,avalle2024persistent}, while \citet{ferron-etal-2023-meep}
define subdimensions such as response diversity, interestingness, and specificity. 



\paragraph{Persuasion}

Empirical literature has primarily examined factors influencing persuasion that align with other categories in our taxonomy, such as linguistic markers (\S\ref{subsec:structure_logic}) and turn-taking 
(\S\ref{subsec:social_dynamics}) \cite{10.1145/2872427.2883081}. Considering this connection, persuasion is not only an indicator of argument quality, but may also serve as a proxy for identifying additional markers signaling whether facilitator intervention is needed.





\paragraph{Diversity and Informativeness} Diversity in online discussions refers to the presence of varied perspectives, backgrounds, and experiences, which can enrich conversations by fostering constructive exchanges \cite{irani2024argusense,zhang2024comprehensive}. To prevent echo chambers and promote inclusivity, facilitators can use diversity measures to encourage opinion diversity \cite{Anastasiou23}, encouraging users to explore a broad range of perspectives on a given issue \cite{kim_et_al_chatbot}. Informativeness refers to the relevance and value of information shared in a discussion and is considered a building stock of prosociality, an attribute that facilitation trys to foster in online discussion platforms \cite{Lambert2024}. 






\subsection{LLM Approaches to Discussion Quality}

\acp{LLM} can significantly aid in evaluating discussion quality, performing on par with humans in annotating argument structure \cite{mirzakhmedova2024large, rescala-etal-2024-language}, excelling in comparative argument evaluation \cite{wang2023contextual}, \ac{AM}, and synthesis \cite{chen2024exploring, irani2024argusense, anastasiou2024hybrid}. They are increasingly used  for coherence evaluation at the comment or whole discussion level \cite{zhang2024comprehensive}, often using proprietary models (e.g., GPT-4), while fine-tuned open-source models also show promise \cite{mendonca-etal-2024-ecoh, zhang-etal-2023-xdial}. \acp{LLM} are not preferred for turn-taking or linguistic markers. Research on the former focuses on visual dashboards (such as VisArgue or TurnViz) that reveal dominance shifts at a glance \citep{el-assady-etal-2017-interactive,10.1145/2856767.2856782}, while distinguishing linguistic markers is often approached through older methodologies such as LSTMs \cite{Sak2014LongSM} or dictionaries, as mentioned in \S\ref{subsec:structure_logic}.

\acp{LLM} can also serve as dialogue and speech act annotators \cite{ziems2024can, cimino-etal-2024-coherence, martinenghi-etal-2024-llms, schroeder-etal-2024-fora}. For example, \citet{yu2024} show that GPT-4 reached almost human accuracy on the task of annotating the speech act of apologizing. However, we acknowledge that the difficulty of automatic speech act annotation might depend on the task and more research on that is encouraged. 

Remaining on the frontier of pragmatics, research shows that \ac{LLM}-fine-tuning enhances implicature comprehension \cite{10.5555/3666122.3667035}, with GPT-4 achieving human-level performance through chain-of-thought prompting. While \acp{LLM} perform well in some pragmatics tasks, such as in the Pragmatic Understanding Benchmark (PUB) \cite{sravanthi-etal-2024-pub}, they struggle with social norm-based understanding (e.g., humor, irony) \cite{hu-etal-2023-fine,sravanthi-etal-2024-pub}. This is also true for annotating politeness, power, disagreement, and toxicity \cite{zhou-etal-2024-llm-feature, ziems2024can}. 

\acp{LLM} perform well in identifying power differentials in discussions \cite{ziems2024can}, and can detect these imbalances in real time, enabling facilitators to invite quieter voices and limit dominant turns. Additionally, \acp{LLM} have been successfully employed as dispute tactics annotators, highlighting instances of hostile interactions that may require moderator intervention \cite{zhou-etal-2024-llm-feature}. However, they show limited accuracy in sentiment and engagement detection \cite{hu-etal-2023-fine, sravanthi-etal-2024-pub, furniturewala-jaidka-2024-empaths, xu-jiang-2024-multi}. Empathy detection also remains challenging for \acp{LLM}, with evaluations showing inconsistent performance across conversational tasks \cite{furniturewala-jaidka-2024-empaths, xu-jiang-2024-multi, ziems2024can}. While \acp{LLM} show promise in measuring controversy and persuasion, performance drops at the discussion level, particularly when assessing diversity, informativeness, and broader aspects of sociopragmatic understanding \cite{ziems2024can, avalle2024persistent, 10.1162/coli_a_00364}.

\section{Intervention Strategies}\label{sec:intervention_strategies}

\subsection{When to Intervene}

Picking the right moment to intervene is a crucial part of effective facilitation strategies. If a facilitator does not intervene when they should have, there is a risk of significant escalation, while intervening when unnecessary can increase toxicity \cite{schaffner_community_guidelines, make_reddit_great, proactive_moderation, cresci_pesonalized_interventions}. Even `softer' interventions such as information and opinion sharing can prove detrimental to discussions when performed excessively \cite{gao-etal-2025-moderation}. It is imperative then for a facilitator to be able to recognize subtle cues that hint towards escalation (also considering the evaluation dimensions discussed in \S\ref{sec:eval_methods}), in order to defuse the situation, something that even experienced human facilitators are not confident to reliably do \cite{proactive_moderation}.

The \ac{NLP} task of `Conversational Forecasting' may 
contribute towards this direction. Given a conversation up to a point, a model attempts to predict if an event will occur in the future in that conversation. In our case, this is where 
a facilitator would intervene \cite{proactive_moderation}. 
Traditional \ac{ML} models can perform well on this task, although their performance varies \cite{falk-etal-2021-predicting, park_et_al_2012_facilitation, falk-etal-2024-moderation, proactive_moderation}.

\subsection{How to Intervene}

There is currently no agreed-upon taxonomy for 
facilitator interventions. \citet{cheung-et-al-2011} propose a taxonomy that  focuses on discussion facilitation, excluding, however, disciplinary or administrative actions, which are common in online discussions. \citet{park_et_al_2012_facilitation} propose another taxonomy consisting of seven moderator functions, ranging from policing the discussion to solving technical issues. Their taxonomy, however, is not easily generalizable to domains other than website facilitation \cite{chen-etal-2025-whow}.
These functions roughly correlate with the volunteer moderator roles, as described by \citet{seering_self_moderation}. More practical approaches can be found in facilitator manuals \cite{Cornell_eRulemaking2017, dimitra-guide} and  books \cite{dimitra-book}. \citet{chen-etal-2025-whow} bridge the questions of \emph{when} and \emph{how} to facilitate by proposing a taxonomy that analyzes both individually, which was improved by \citet{gao-etal-2025-moderation}.

Several works have examined facilitation in education. \citet{SJOLIE2021103477} conducted a mixed-methods study on a meta-facilitative approach, where students and teachers explicitly discussed their collaboration and which led to significant learning improvements. In the context of virtual facilitation, \citet{Verkuyl2024} showed that successful integration of virtual simulations in higher education depends not just on access, but on facilitators who align simulations with course objectives, respond to learners’ needs, and evaluate the experience. Both studies suggest that facilitation requires socially informed practices, even as automation promises workload reduction. 

With reference to \ac{NLP} approaches in facilitation in education, \citet{lugini-etal-2020-discussion} designed Discussion Tracker, a classroom analytics tool that applies algorithms to identify argumentation moves (claim, evidence, explanation) and evaluate levels of specificity, as well as recognize patterns of collaboration. Deployment in class showed that teachers considered the analytics valuable, and that the system’s classifiers achieved moderate to substantial agreement with human judgments. The aforementioned work of \citet{gao-etal-2025-moderation} presented an approach that combines automatic \ac{ESL} dialogue assessment with a framework of moderation strategies. The authors showed that moderators improve topic flow and conversation management, with active acknowledgment and encouragement proving most effective, but excessive input can hinder discussion.

Facilitators often have to decide what form of coercive measure to take to make sure the conversation remains healthy, without having to intervene repeatedly. Human interventions typically use an unofficial `escalation ladder' (Figure~\ref{fig:survey}), where the facilitator will progressively move from milder facilitation tactics to threatening, and finally enacting disciplinary action \cite{seering_self_moderation}. `Conversational moderation' \cite{cho-etal-2024-language}, where a facilitator first converses with the offender, has proven effective and is actively encouraged in some facilitator guidelines \cite{thecommons2025}. This is probably why disciplinary action is typically not the first choice of a facilitator \cite{proactive_moderation} and why it should reasonably be used as a last resort.

Softer kinds of interventions that facilitators frequently use first include: setting and informing users about rules \cite{proactive_moderation, seering_self_moderation}, welcoming new users \cite{proactive_moderation}, summarizing key points \cite{small-polis-llm, falk-etal-2024-moderation}, balancing participation \cite{kim_et_al_chatbot, fishkin2018deliberative}, and aiding users improve their points \cite{Tsai2024Generative, falk-etal-2024-moderation}. Facilitators are also instrumental in beginning and ending discussions \cite{small-polis-llm, gao-etal-2025-moderation}, as well as generally encouraging participants \cite{gao-etal-2025-moderation}. It is worth noting that facilitation guides may explicitly forbid facilitators from intervening in certain ways, such as sharing their opinions or providing new information \cite{dimitra-guide}. 

\subsection{Personalized Interventions}

Intervention strategies should not be applied en masse, without considering the characteristics of each individual. Traditionally, massive 
application of disciplinary action (or threatening)  has led to adverse effects community- and platform-wide \cite{make_reddit_great, falk-etal-2021-predicting} and to the creation of echo-chambers \cite{cho-etal-2024-language}. There are also calls for research to move away from one-size-fits-all approaches and instead move towards personalized interventions \cite{cresci_pesonalized_interventions}. Human facilitators are often able to personalize interventions per individual \cite{proactive_moderation}, and we hypothesize that \acp{LLM} can also do so to some extent.

\section{Towards LLM-based facilitation} \label{sec:towards_llm_facilitation}


Until recently, \ac{ML} models used as facilitation agents were confined to either performing menial tasks, such as pasting automated messages \cite{seering_self_moderation, proactive_moderation}, suggesting facilitation actions (e.g., rejecting posts), possibly via human-in-the-loop frameworks \cite{fishkin2018deliberative, gelauff_achieving_parity}, identifying possibly escalatory comments \cite{proactive_moderation}, or employing pre-programmed facilitative tactics, as in the work of \citet{kim_et_al_chatbot}, where the model produces automated messages encouraging participation. However, older \ac{ML}-based and rule-based facilitation are not effective enough to meet the high demands of most platforms \cite{seering_self_moderation, schaffner_community_guidelines}.

\begin{figure}
    \centering
    \includegraphics[width=\linewidth]{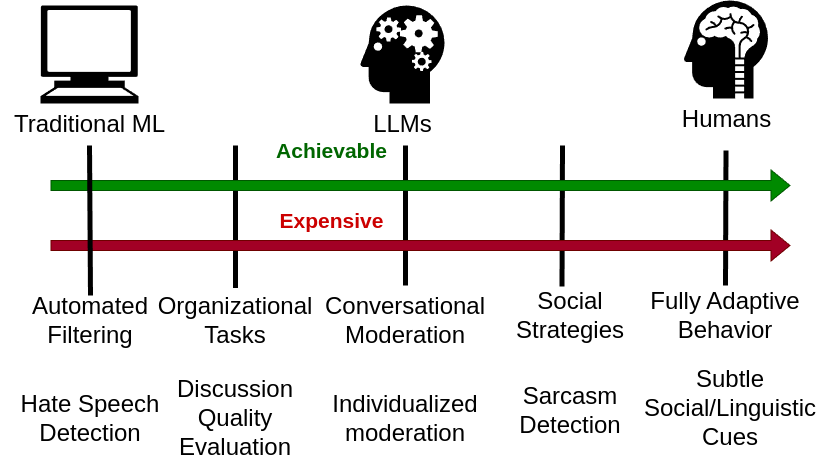}
    \caption{Capabilities of simpler \ac{ML}, \ac{LLM}, and human facilitation. Task complexity and cost increase from left to right. Intermediate tasks are handled suboptimally by the preceding method.}
    \label{fig:intervention_taxonomy}
\end{figure}

Advances in \acp{LLM} enable the development of \emph{facilitation agents} that engage more actively in discussions. These agents can warn users about policy violations \cite{Kumar_AbuHashem_Durumeric_2024}, suggest rephrasings to improve tone or persuasiveness \cite{bose-etal-2023-detoxifying}, monitor turn-taking \cite{schroeder-etal-2024-fora}, and summarize or visualize key discussion points \cite{small-polis-llm}. They can also assist in drafting group statements that reflect diverse viewpoints \cite{Tessler24}. A brief, non-exhaustive summary of the capabilities of simpler \ac{ML} models, \acp{LLM}, and humans can be found in Figure~\ref{fig:intervention_taxonomy}.

\subsection{Administrating the Discussion}

\acp{LLM} are able to tackle a variety of `administrative' facilitation tasks that help structure discussions. For example, facilitators often summarize
the views of the participants, seek confirmation of understanding, and share perspectives. This iterative summarization is a task \acp{LLM} may handle effectively \cite{small-polis-llm, burton2024large}. However, \citet{Feng22} point out some challenges such as discussions with multiple participants, topic drifts, multiple co-references, diverse interactive signals, and diverse domain terminologies. Still, according to \citet{Jin2024}, \acp{LLM} bring significant advantages over conventional \ac{ML} methods, ``notably in the quality and flexibility of the generated texts and the prompting paradigm to alleviate the cost of training deep models''. 

In some deliberative contexts, facilitators are also encouraged to begin a discussion with their own opinion \cite{small-polis-llm}, although others disagree \cite{dimitra-guide}. This is a task \acp{LLM} can also handle, albeit less convincingly than \ac{IR} approaches \cite{karadzhov2023delidata}. 

Finally, \acp{LLM} can help marginalized groups in discussions by offering translations of the discussions in their native languages, and by helping them phrase their opinions with proper grammar and syntax \cite{Tsai2024Generative, burton2024large}. This can directly improve discussions by increasing their diversity (Section~\ref{subsec:engagement_impact}).

\subsection{Evolving Traditional Automation Models}
\label{ssec:llm-mod:ml}

\acp{LLM} have been proven to be adept at \ac{NLP} tasks such as the detection of hate speech \cite{shi-2024-hatespeech}, toxicity \cite{kang-qian-2024-implanting, Wang2022ToxicityDW}, and misinformation \cite{kang-qian-2024-implanting, Wang2022ToxicityDW}. These abilities make \acp{LLM} usable as drop-in replacements for traditional \ac{ML} models for these tasks, suggesting that conversational \ac{LLM} facilitation agents may be able to identify, and dynamically adapt to such phenomena properly. We note however that \acp{LLM} are much more expensive and less scalable than their simpler \ac{ML} counterparts. Furthermore, \ac{LLM} annotation has its own challenges: \ac{LLM} survey responses \cite{jansen_2023, bisbee_2023, neumann_2025} and annotations \cite{Gligoric2024CanUL} are generally unreliable and surface-level. Non-deterministic behavior is also common in 
\acp{LLM} \cite{atil_2025}, but also particularly in closed-source models \cite{bisbee_2023} on which a lot of research on \ac{LLM} annotation hinges.

\subsection{Fully Automated LLM-based Facilitation}

There are indications that
\acp{LLM} can be used as facilitators to the fullest capacity of the role. \acp{LLM} are able to predict optimal facilitation tactics \cite{schroeder-etal-2024-fora}, like traditional \ac{ML} models \cite{al-khatib-etal-2018-modeling}. Furthermore, they have proven capable of developing and executing social strategies in other tasks, e.g., negotation games, \ac{LLM} interactions \cite{Abdelnabi2023CooperationCA, Cheng2024SelfplayingAL, martinenghi-etal-2024-llms}. Given that relatively simple \ac{ML} chatbots, which do not leverage generative text capabilities, have been 
reported to improve discussions \cite{kim_et_al_chatbot}, many expect \ac{LLM}-based facilitation to be a promising solution to the well-known bottleneck of human facilitation \cite{small-polis-llm, seering_self_moderation, burton2024large, schroeder-etal-2024-fora}. Notably, \citet{cho-etal-2024-language} successfully use \ac{LLM} facilitators with prompts based on Cognitive Behavioral Therapy 
to moderate a live discussion with human participants. Their work shows that \ac{LLM} facilitators can adapt their instructions to users, although they cannot by themselves affect the discussion with regard to cooperation and mutual respect between the participants.

Nevertheless, \acp{LLM} have inherent limitations that make them worse than humans in most social tasks (Figure~\ref{fig:intervention_taxonomy}; \citet{rossi_2024}). While human facilitators are encouraged to be neutral \cite{dimitra-book, Cornell_eRulemaking2017}, numerous studies point to biases in sociodemographic, statistical, and political terms in \acp{LLM} \cite{anthis_2025, hewitt2024predicting, rossi_2024}, which can be exacerbated during the course of a discussion \cite{Taubenfeld2024SystematicBI}. 



\section{Facilitation Datasets}
\label{sec:datasets}

In this section, we provide an overview of the most prominent datasets for online facilitation, considering their sizes and 
their relevance to core facilitation tasks. These datasets can be used for analyzing the behavior of human facilitators and the reactions of the participants, investigating the existing taxonomies (e.g., ones presented in \S\ref{sec:intervention_strategies}) or as training data for human and \ac{LLM} facilitators.

Due to the low number of such datasets in literature \cite{chen-etal-2025-whow}, the entries presented in this section straddle various domains adjacent to online facilitation. Hence, we propose the following new taxonomy of facilitation datasets: \emph{Conversation Derailment} datasets, where the task is to predict when a conversation escalates, therefore requiring facilitator intervention; and \emph{Facilitator Interventions} datasets, which include comments by facilitators in active discussions, sometimes annotated with the tactics employed. Some datasets contain information that can be used in multiple tasks.\footnote{Despite its designation, the `WikiDisputes' dataset does include information about facilitators. We consider it solely a `Conversation Derailment' dataset because facilitator interventions only constitute $0.03\%$ of its comments.} An overview of the surveyed  datasets and their categories in our taxonomy can be found in Table~\ref{tab::datasets}.

\begin{table*}[!t]
\centering
{ 
\fontsize{7pt}{9pt}\selectfont 
\renewcommand{\arraystretch}{0.95} 
\setlength{\tabcolsep}{2.5pt} 
\begin{adjustbox}{width=\textwidth}
\begin{tabular}
    { |>{\raggedright}p{3.2cm}|p{1.6cm}|p{1.6cm}|p{1.5cm}|>{\raggedright\arraybackslash}p{5.2cm}| }
    \hline
    \cellcolor{blue!25}\textbf{Name} & \multicolumn{2}{|c|}{\cellcolor{blue!25}\textbf{Task}} & \cellcolor{blue!25}\textbf{Size} & \cellcolor{blue!25}\textbf{Content} \\
    \hline
    WikiDisputes \cite{de-kock-vlachos-2021-beg} & \multicolumn{2}{|c|}{\taskderailment} & $7{,}425$ D & Includes annotations for several `dispute tactics.' \\
    \hline
    Wiki-Tactics \cite{de-kock-vlachos-2022-disagree} & \taskderailment & \taskfacilitation & $213$ D & Based on Wikipedia Disputes, includes moderation action metadata such as comment edits and deletions.\\
    \hline
    WikiConv \cite{hua2018wikiconvcorpuscompleteconversational} & \multicolumn{2}{|c|}{\taskfacilitation} & $91{,}000{,}000$ D & Includes moderation meta-data such as comment edits and deletions.\\
    \hline
    Conversations Gone Awry \cite{zhang-etal-2018-conversations} & \multicolumn{2}{|c|}{\taskderailment} & $4{,}188$ D & Predicts derailment by analyzing rhetorical tactics, human-annotated.\\
    \hline
    \citet{chang-danescu-niculescu-mizil-2019-trouble} (1) & \multicolumn{2}{|c|}{\taskderailment} & $4{,}188$ D & Extends the `Conversations Gone Awry' dataset. \\
    \hline
    \citet{chang-danescu-niculescu-mizil-2019-trouble} (2) & \multicolumn{2}{|c|}{\taskderailment} & $6{,}842$ D & Based on the r/ChangeMyView subreddit. \\
    \hline
    \citet{park_et_al_2012_facilitation} & \taskderailment & \taskfacilitation & $1{,}678$ C & Comprised of 4 datasets. Includes 19 intervention types belonging to 7 moderator roles.\\
    \hline
    RegulationRoom \cite{falk-etal-2021-predicting} & \multicolumn{2}{|c|}{\taskderailment} & $3{,}000$ C & Extends the dataset of \citet{park_et_al_2012_facilitation}.\\
    \hline
    UMOD \cite{falk-etal-2024-moderation} & \multicolumn{2}{|c|}{\taskfacilitation} & $2{,}000$ C & Based on the r/ChangeMyView subreddit, annotated for facilitation tactics and \ac{AQ}.\\
    \hline
    Fora \cite{schroeder-etal-2024-fora} & \multicolumn{2}{|c|}{\taskfacilitation} & $262$ D & Original dataset revolving around experience-sharing, annotated for facilitation tactics.\\
    \hline
    WHoW \cite{chen-etal-2025-whow} & \multicolumn{2}{|c|}{\taskfacilitation} & $21{,}151$ C & Dataset derived from TV debates and radio panels, annotated for facilitation tactics.\\
    \hline
   L2Moderator \cite{gao-etal-2025-moderation} & \multicolumn{2}{|c|}{\taskfacilitation} & 17 D (16.5 hours of transcripts) & Facilitated online discussions for \ac{ESL} speakers.\\
    \hline
\end{tabular}
\end{adjustbox}
} 
\caption{Overview of reviewed datasets. Unnamed datasets are referred to by the names of the authors only. The size reflects the number of annotated conversations, disregarding unlabeled data. 
\textbf{D} indicates the number of discussions. \textbf{C} indicates the number of individual comments or dialogue turns.}
\label{tab::datasets}
\end{table*}

\section{LLM Discussion Facilitation Roadmap}
\paragraph{Evaluation}
\acp{LLM} can serve as automated discussion quality annotators (\S\ref{sec:eval_methods}). Are these annotators infallible? Not yet. Certain dimensions, especially those that are highly subjective (e.g., pragmatic understanding), remain challenging for \acp{LLM} to annotate accurately. But we must take into account that even human annotations tend to be polarized for such subjective quality dimensions \cite{argyle2023}, largely due to sociodemographic background effects and personal biases \cite{beck-etal-2024-sensitivity, sap-etal-2020-social}. 


On the other hand, prompted \acp{LLM} offer a more scalable and cost-effective alternative for annotating discussion quality compared to human annotation and traditional (or self-) supervised training on large annotated datasets.  
Using \acp{LLM} for annotation, however, requires careful model selection considering whether models are open or closed source, model size, model alignment, 
as well as prompt selection, and (if applicable) fine-tuning requirements. These choices should be tailored to the specific quality dimension being evaluated.


\paragraph{Facilitation}
Intervention types should be adapted to the different legal frameworks, rules, and social norms of each community/platform. While there are exhaustive surveys on intervention types and policies, such as that of \citet{schaffner_community_guidelines}, there is yet no methodology to train human or 
artificial facilitators according to these factors. We posit that experiments using exclusively \ac{LLM} user/facilitator-agents are necessary to sustainably test new facilitation strategies and interventions per community and platform, as 
in other \ac{NLP} tasks that involve \ac{LLM}-generated conversation \cite{ulmer2024bootstrappingllmbasedtaskorienteddialogue, cheng2024selfplayingadversariallanguagegame, park2022socialsimulacracreatingpopulated, Park2023GenerativeAI}, before testing the resulting facilitators in costly experiments with human participants. Finally, the datasets presented in Table \ref{tab::datasets} can be used to train and assess \ac{LLM} facilitators in the future, as well as to generate additional data—similar to the existing ones, but with controlled modifications—to stress-test various 
facilitators in particular settings (e.g., predicting or recovering from a conversation derailment).


\section{Conclusions}


This survey examined online discussion evaluation and facilitation by bridging insights from Social Science and \ac{NLP}, with a focus on the growing role of \acp{LLM}. We introduced a new discussion evaluation taxonomy, with categories that should remain flexible depending on the evaluation task and the characteristics of the discussion.
In terms of intervention strategies, both human- and machine-driven advancements show significant promise in improving the quality of interventions, helping online discussions remain constructive, and resistant to derailment.  Most facilitation datasets still originate from human online conversations, with research yet to fully explore the capabilities of \acp{LLM}. Taking the above into account, we believe that now is the time to embrace \acp{LLM} for facilitation to foster healthier and more constructive conversations.

\section{Limitations}

This survey is not without its limitations. While we have attempted to present a comprehensive overview of facilitation methods, certain techniques, such as summarization, could be explored in greater depth. Since summarization is a vast subfield of NLP, it was only briefly mentioned.

Moreover, it is important to highlight that most research on facilitation has been conducted solely in English-speaking online spaces. The inherent limitations of \acp{LLM} in handling other languages and cultural contexts must be considered. As a result, these findings may not be easily applicable to other regions of the world.

Finally, the majority of  real-world online discussions and 
deliberations happen in the context of communities, where group dynamics (social behaviors, power structures, norms, and interactions) apply. Thus, a fuller review of facilitation would have to account for the internal dynamics of such communities, as well as the wider role of the facilitator as a figure that not only helps in the conversation but has a social status in the group as well. 


\section{Ethical Considerations}
Although AI, and  \acp{LLM} in particular, can be effectively used as discussion facilitators, offering dynamic, responsive discussion support, their deployment must meet strict transparency, safety, and accountability standards, especially for high-risk applications, as stated in the EU AI Act.\footnote{\url{https://digital-strategy.ec.europa.eu/en/policies/regulatory-framework-ai}}
For example, a person or minority group may have been unfairly disadvantaged in an AI-enhanced  deliberation.
It is also necessary for the users to be aware that they are interacting with AI facilitators. Ideally, the consent of the users should be sought before  using any sort of AI-enhanced discussion platform.

Even if \acp{LLM} facilitators eventually achieve a high level of autonomy, it is advisable to maintain human oversight. Keeping a human-in-the-loop approach ensures greater transparency and enables effective error prevention, detection, and correction.


\section*{Acknowledgments}
This work has been partially supported by project MIS 5154714 of the National Recovery and Resilience Plan Greece 2.0 funded by the European Union under the NextGenerationEU Program.

\bibliography{acl_latex}

\begin{thebibliography}{155}
\providecommand{\natexlab}[1]{#1}

\bibitem[{Abdelnabi et~al.(2024)Abdelnabi, Gomaa, Sivaprasad, Schönherr, and Fritz}]{Abdelnabi2023CooperationCA}
S.~Abdelnabi, A.~Gomaa, S.~Sivaprasad, L.~Schönherr, and M.~Fritz. 2024.
\newblock Cooperation, competition, and maliciousness: {LLM}-stakeholders interactive negotiation.
\newblock In \emph{The Thirty-eight Conference on Neural Information Processing Systems Datasets and Benchmarks Track}, pages 96--106, Vancouver, Canada.

\bibitem[{Adomavicius(2021)}]{adomavicius2021}
S.~Adomavicius. 2021.
\newblock \href {https://docs.rwu.edu/nyscaproceedings/vol2017/iss1/12} {Putting the social in social media: How human connection triggers engagement}.
\newblock In \emph{Proceedings of the New York State Communication Association}, volume 2017.

\bibitem[{Aher et~al.(2023)Aher, Arriaga., and Kalai}]{Aher2023}
G.~Aher, R.I. Arriaga., and A.T. Kalai. 2023.
\newblock Using large language models to simulate multiple humans and replicate human subject studies.
\newblock In \emph{Proceedings of the 40th International Conference on Machine Learning}, pages 337 -- 371, Hawaii, USA.

\bibitem[{Al-Khatib et~al.(2018)Al-Khatib, Wachsmuth, Lang, Herpel, Hagen, and Stein}]{al-khatib-etal-2018-modeling}
K.~Al-Khatib, H.~Wachsmuth, K.~Lang, J.~Herpel, M.~Hagen, and B.~Stein. 2018.
\newblock \href {https://doi.org/10.18653/v1/P18-1237} {Modeling deliberative argumentation strategies on {W}ikipedia}.
\newblock In \emph{Proceedings of the 56th Annual Meeting of the Association for Computational Linguistics (Volume 1: Long Papers)}, pages 2545--2555, Melbourne, Australia.

\bibitem[{Anastasiou and De~Liddo(2024)}]{anastasiou2024hybrid}
L.~Anastasiou and A.~De~Liddo. 2024.
\newblock A hybrid human-{AI} approach for argument map creation from transcripts.
\newblock In \emph{Proceedings of the First Workshop on Language-driven Deliberation Technology (DELITE)@ LREC-COLING 2024}, pages 45--51, Turin, Italy.

\bibitem[{Anastasiou et~al.(2023)Anastasiou, De~Moor, Brayshay, and De~Liddo}]{Anastasiou23}
L.~Anastasiou, A.~De~Moor, B.~Brayshay, and A.~De~Liddo. 2023.
\newblock \href {https://doi.org/10.1145/3593743.3593771} {A tale of struggles: an evaluation framework for transitioning from individually usable to community-useful online deliberation tools}.
\newblock In \emph{Proceedings of the 11th International Conference on Communities and Technologies}, C\&T '23, page 144–155, New York, NY, USA. Association for Computing Machinery.

\bibitem[{Anthis et~al.(2025)Anthis, L., Richardson, Kozlowski, Koch, Evans, Brynjolfsson, and Bernstein}]{anthis_2025}
J.~R. Anthis, R.~L., S.~M. Richardson, A.~C. Kozlowski, B.~Koch, J.~Evans, E.~Brynjolfsson, and M.~Bernstein. 2025.
\newblock \href {https://arxiv.org/abs/2504.02234} {Llm social simulations are a promising research method}.
\newblock \emph{Preprint}, arXiv:2504.02234.

\bibitem[{Argyle et~al.(2023)Argyle, Bail, Busby, Gubler, Howe, Rytting, Sorensen, and Wingate}]{argyle2023}
L.~P. Argyle, C.~A. Bail, E.~C. Busby, J.~R. Gubler, T.~Howe, C.~Rytting, T.~Sorensen, and D.~Wingate. 2023.
\newblock Leveraging {AI} for democratic discourse: Chat interventions can improve online political conversations at scale.
\newblock \emph{Proceedings of the National Academy of Sciences}, 120(41):1--8.

\bibitem[{Asterhan and Schwarz(2010)}]{Asterhan2010OnlineMO}
C.~S.~C. Asterhan and B.~B. Schwarz. 2010.
\newblock \href {https://api.semanticscholar.org/CorpusID:7857360} {Online moderation of synchronous e-argumentation}.
\newblock \emph{International Journal of Computer-Supported Collaborative Learning}, 5:259--282.

\bibitem[{Atil et~al.(2025)Atil, Aykent, Chittams, Fu, Passonneau, Radcliffe, Rajagopal, Sloan, Tudrej, Ture, Wu, Xu, and Baldwin}]{atil_2025}
B.~Atil, S.~Aykent, A.~Chittams, L.~Fu, R.~J. Passonneau, E.~Radcliffe, G.~R. Rajagopal, A.~Sloan, T.~Tudrej, F.~Ture, Z.~Wu, L.~Xu, and B.~Baldwin. 2025.
\newblock \href {https://arxiv.org/abs/2408.04667} {Non-determinism of "deterministic" llm settings}.
\newblock \emph{Preprint}, arXiv:2408.04667.

\bibitem[{Austin(1975)}]{austin1975things}
J.~L. Austin. 1975.
\newblock \emph{How to Do Things with Words}.
\newblock Oxford University Press.

\bibitem[{Avalle et~al.(2024)Avalle, Di~Marco, Etta, Sangiorgio, Alipour, Bonetti, Alvisi, Scala, Baronchelli, Cinelli et~al.}]{avalle2024persistent}
M.~Avalle, N.~Di~Marco, G.~Etta, E.~Sangiorgio, S.~Alipour, A.~Bonetti, L.~Alvisi, A.~Scala, A.~Baronchelli, M.~Cinelli, et~al. 2024.
\newblock Persistent interaction patterns across social media platforms and over time.
\newblock \emph{Nature}, 628(8008):582--589.

\bibitem[{Bang et~al.(2023)Bang, Yu, Madotto, Lin, Diab, and Fung}]{bang-etal-2023-enabling}
Y.~Bang, T.~Yu, A.~Madotto, Z.~Lin, M.~Diab, and P.~Fung. 2023.
\newblock Enabling classifiers to make judgements explicitly aligned with human values.
\newblock In \emph{Proceedings of the 3rd Workshop on Trustworthy {NLP}}, pages 311--325, Toronto, Canada.

\bibitem[{Bawden(2021)}]{bawden-2021-understanding}
R.~Bawden. 2021.
\newblock Understanding dialogue: Language use and social interaction.
\newblock \emph{Computational Linguistics}, 47(3):703--705.

\bibitem[{Beck et~al.(2024)Beck, Schuff, Lauscher, and Gurevych}]{beck-etal-2024-sensitivity}
T.~Beck, H.~Schuff, A.~Lauscher, and I.~Gurevych. 2024.
\newblock Sensitivity, performance, robustness: Deconstructing the effect of sociodemographic prompting.
\newblock In \emph{Proceedings of the 18th Conference of the European Chapter of the Association for Computational Linguistics (Volume 1: Long Papers)}, pages 2589--2615, Malta.

\bibitem[{Benesch et~al.(2016)Benesch, Ruths, Dillon, Saleem, and Wright}]{benesch2016counterspeech}
S.~Benesch, D.~Ruths, K.~P. Dillon, H.~M. Saleem, and L.~Wright. 2016.
\newblock Counterspeech on twitter: A field study. dangerous speech project.

\bibitem[{Bisbee et~al.(2024)Bisbee, Clinton, Dorff, Kenkel, and Larson}]{bisbee_2023}
J.~Bisbee, J.~D. Clinton, C.~Dorff, B.~Kenkel, and J.~M. Larson. 2024.
\newblock \href {https://doi.org/10.1017/pan.2024.5} {Synthetic replacements for human survey data? the perils of large language models}.
\newblock \emph{Political Analysis}, 32(4):401–416.

\bibitem[{Bose et~al.(2023)Bose, Perera, and Dorr}]{bose-etal-2023-detoxifying}
R.~Bose, I.~Perera, and B.~Dorr. 2023.
\newblock Detoxifying online discourse: A guided response generation approach for reducing toxicity in user-generated text.
\newblock In \emph{Proceedings of the First Workshop on Social Influence in Conversations}, pages 9--14, Toronto, Canada.

\bibitem[{Burton et~al.(2024)Burton, Lopez-Lopez, Hechtlinger et~al.}]{burton2024large}
J.~W. Burton, E.~Lopez-Lopez, S.~Hechtlinger, et~al. 2024.
\newblock How large language models can reshape collective intelligence.
\newblock \emph{Nature Human Behaviour}, 8:1643--1655.

\bibitem[{Bächtiger et~al.(2022)Bächtiger, Gerber, and Fournier-Tombs}]{10.1093/oso/9780192848925.003.0006}
A.~Bächtiger, M.~Gerber, and E.~Fournier-Tombs. 2022.
\newblock \href {https://doi.org/10.1093/oso/9780192848925.003.0006} {83discourse quality index}.
\newblock In \emph{Research Methods in Deliberative Democracy}. Oxford University Press.

\bibitem[{Bächtiger et~al.(2010)Bächtiger, Niemeyer, Neblo, Steenbergen, and Steiner}]{https://doi.org/10.1111/j.1467-9760.2009.00342.x}
A.~Bächtiger, S.~Niemeyer, M.~Neblo, M.~R. Steenbergen, and J.~Steiner. 2010.
\newblock \href {https://doi.org/10.1111/j.1467-9760.2009.00342.x} {Disentangling diversity in deliberative democracy: Competing theories, their blind spots and complementarities}.
\newblock \emph{Journal of Political Philosophy}, 18(1):32--63.

\bibitem[{Cervone and Riccardi(2020)}]{cervone-riccardi-2020-dialogue}
A.~Cervone and G.~Riccardi. 2020.
\newblock \href {https://doi.org/10.18653/v1/2020.sigdial-1.21} {Is this dialogue coherent? learning from dialogue acts and entities}.
\newblock In \emph{Proceedings of the 21th Annual Meeting of the Special Interest Group on Discourse and Dialogue}, pages 162--174, online. Association for Computational Linguistics.

\bibitem[{Chang and Danescu-Niculescu-Mizil(2019)}]{chang-danescu-niculescu-mizil-2019-trouble}
J.~P. Chang and C.~Danescu-Niculescu-Mizil. 2019.
\newblock \href {https://doi.org/10.18653/v1/D19-1481} {Trouble on the horizon: Forecasting the derailment of online conversations as they develop}.
\newblock In \emph{Proceedings of the 2019 Conference on Empirical Methods in Natural Language Processing and the 9th International Joint Conference on Natural Language Processing (EMNLP-IJCNLP)}, pages 4743--4754, Hong Kong, China. Association for Computational Linguistics.

\bibitem[{Chen et~al.(2024)Chen, Cheng, Tuan, and Bing}]{chen2024exploring}
G.~Chen, L.~Cheng, L.~A. Tuan, and L.~Bing. 2024.
\newblock Exploring the potential of large language models in computational argumentation.
\newblock In \emph{Proceedings of the 62nd Annual Meeting of the Association for Computational Linguistics (Volume 1: Long Papers)}, pages 2309--2330.

\bibitem[{Chen et~al.(2025)Chen, Frermann, and Lau}]{chen-etal-2025-whow}
M.~Chen, L.~Frermann, and J.~H. Lau. 2025.
\newblock \href {https://doi.org/10.18653/v1/2025.naacl-long.105} {{WH}o{W}: A cross-domain approach for analysing conversation moderation}.
\newblock In \emph{Proceedings of the 2025 Conference of the Nations of the Americas Chapter of the Association for Computational Linguistics: Human Language Technologies (Volume 1: Long Papers)}, pages 2091--2126, Albuquerque, New Mexico. Association for Computational Linguistics.

\bibitem[{Cheng et~al.(2024{\natexlab{a}})Cheng, Hu, Xu, Zhang, Dai, Han, and Du}]{Cheng2024SelfplayingAL}
P.~Cheng, T.~Hu, H.~Xu, Z.~Zhang, Y.~Dai, L.~Han, and N.~Du. 2024{\natexlab{a}}.
\newblock \href {https://api.semanticscholar.org/CorpusID:269157364} {Self-playing adversarial language game enhances llm reasoning}.
\newblock \emph{ArXiv}, abs/2404.10642.

\bibitem[{Cheng et~al.(2024{\natexlab{b}})Cheng, Hu, Xu, Zhang, Dai, Han, and Du}]{cheng2024selfplayingadversariallanguagegame}
P.~Cheng, T.~Hu, H.~Xu, Z.~Zhang, Y.~Dai, L.~Han, and N.~Du. 2024{\natexlab{b}}.
\newblock \href {https://api.semanticscholar.org/CorpusID:269157364} {Self-playing adversarial language game enhances llm reasoning}.
\newblock \emph{ArXiv}, abs/2404.10642.

\bibitem[{Cho et~al.(2024)Cho, Liu, Shi, Jain, Rizk, Huang, Lu, Wen, Gratch, Ferrara, and May}]{cho-etal-2024-language}
H.~Cho, S.~Liu, T.~Shi, D.~Jain, B.~Rizk, Y.~Huang, Z.~Lu, N.~Wen, J.~Gratch, E.~Ferrara, and J.~May. 2024.
\newblock Can language model moderators improve the health of online discourse?
\newblock In \emph{Proceedings of the 2024 Conference of the North American Chapter of the Association for Computational Linguistics: Human Language Technologies (Volume 1: Long Papers)}, pages 7478--7496, Mexico City, Mexico.

\bibitem[{Cimino et~al.(2024)Cimino, Li, Carenini, and Deufemia}]{cimino-etal-2024-coherence}
G.~Cimino, C.~Li, G.~Carenini, and V.~Deufemia. 2024.
\newblock \href {https://doi.org/10.18653/v1/2024.sigdial-1.26} {Coherence-based dialogue discourse structure extraction using open-source large language models}.
\newblock In \emph{Proceedings of the 25th Annual Meeting of the Special Interest Group on Discourse and Dialogue}, pages 297--316, Kyoto, Japan.

\bibitem[{Concannon and Tomalin(2024)}]{Concannon2024}
S.~Concannon and M.~Tomalin. 2024.
\newblock \href {https://doi.org/10.1007/s00146-023-01715-z} {Measuring perceived empathy in dialogue systems}.
\newblock \emph{AI \& Society}, 39:2233--2247.

\bibitem[{Cresci et~al.(2022)Cresci, Trujillo, and Fagni}]{cresci_pesonalized_interventions}
S.~Cresci, A.~Trujillo, and T.~Fagni. 2022.
\newblock \href {https://doi.org/10.1145/3511095.3536369} {Personalized interventions for online moderation}.
\newblock In \emph{Proceedings of the 33rd ACM Conference on Hypertext and Social Media}, page 248–251, New York, NY, USA.

\bibitem[{Danescu-Niculescu-Mizil et~al.(2012)Danescu-Niculescu-Mizil, Lee, Pang, and Kleinberg}]{10.1145/2187836.2187931}
C.~Danescu-Niculescu-Mizil, L.~Lee, B.~Pang, and J.~Kleinberg. 2012.
\newblock \href {https://doi.org/10.1145/2187836.2187931} {Echoes of power: language effects and power differences in social interaction}.
\newblock In \emph{Proceedings of the 21st International Conference on World Wide Web}, page 699–708, New York, NY, USA.

\bibitem[{De~Kock et~al.(2022)De~Kock, Stafford, and Vlachos}]{de-kock-vlachos-2022-disagree}
C.~De~Kock, T.~Stafford, and A.~Vlachos. 2022.
\newblock How to disagree well: Investigating the dispute tactics used on {W}ikipedia.
\newblock In \emph{Proceedings of the 2022 Conference on Empirical Methods in Natural Language Processing}, pages 3824--3837, Abu Dhabi, United Arab Emirates.

\bibitem[{De~Kock and Vlachos(2021)}]{de-kock-vlachos-2021-beg}
C.~De~Kock and A.~Vlachos. 2021.
\newblock \href {https://doi.org/10.18653/v1/2021.eacl-main.173} {{I} beg to differ: A study of constructive disagreement in online conversations}.
\newblock In \emph{Proceedings of the 16th Conference of the European Chapter of the Association for Computational Linguistics: Main Volume}, pages 2017--2027, Online.

\bibitem[{Degeling et~al.(2015)Degeling, Carter, and Rychetnik}]{DEGELING2015114}
C.~Degeling, S.~M. Carter, and L.~Rychetnik. 2015.
\newblock \href {https://doi.org/10.1016/j.socscimed.2015.03.009} {Which public and why deliberate? – a scoping review of public deliberation in public health and health policy research}.
\newblock \emph{Social Science \& Medicine}, 131:114--121.

\bibitem[{El-Assady et~al.(2017)El-Assady, Hautli-Janisz, Gold, Butt, Holzinger, and Keim}]{el-assady-etal-2017-interactive}
M.~El-Assady, A.~Hautli-Janisz, V.~Gold, M.~Butt, K.~Holzinger, and D.~Keim. 2017.
\newblock \href {https://aclanthology.org/P17-4009/} {Interactive visual analysis of transcribed multi-party discourse}.
\newblock In \emph{Proceedings of {ACL} 2017, System Demonstrations}, pages 49--54, Vancouver, Canada.

\bibitem[{eRulemaking Initiative(2017)}]{Cornell_eRulemaking2017}
Cornell eRulemaking Initiative. 2017.
\newblock \href {https://scholarship.law.cornell.edu/ceri/21} {Ceri (cornell e-rulemaking) moderator protocol}.
\newblock Cornell e-Rulemaking Initiative Publications, 21.

\bibitem[{Falk et~al.(2021)Falk, Jundi, Vecchi, and Lapesa}]{falk-etal-2021-predicting}
N.~Falk, I.~Jundi, E.~M. Vecchi, and G.~Lapesa. 2021.
\newblock \href {https://doi.org/10.18653/v1/2021.argmining-1.13} {Predicting moderation of deliberative arguments: Is argument quality the key?}
\newblock In \emph{Proceedings of the 8th Workshop on Argument Mining}, pages 133--141, Punta Cana, Dominican Republic.

\bibitem[{Falk and Lapesa(2023)}]{falk2023bridging}
N.~Falk and G.~Lapesa. 2023.
\newblock Bridging argument quality and deliberative quality annotations with adapters.
\newblock In \emph{Findings of the Association for Computational Linguistics: EACL 2023}, pages 2469--2488.

\bibitem[{Falk et~al.(2024)Falk, Vecchi, Jundi, and Lapesa}]{falk-etal-2024-moderation}
N.~Falk, E.~Vecchi, I.~Jundi, and G.~Lapesa. 2024.
\newblock \href {https://aclanthology.org/2024.eacl-long.60} {Moderation in the wild: Investigating user-driven moderation in online discussions}.
\newblock In \emph{Proceedings of the 18th Conference of the European Chapter of the Association for Computational Linguistics (Volume 1: Long Papers)}, pages 992--1013, St. Julian{'}s, Malta.

\bibitem[{Feng and Qin(2022)}]{Feng22}
X.~Feng and B.~Qin. 2022.
\newblock \href {https://doi.org/10.24963/ijcai.2022/764} {A survey on dialogue summarization: Recent advances and new frontiers}.
\newblock In \emph{Proceedings of the Thirty-First International Joint Conference on Artificial Intelligence, {IJCAI-22}}, pages 5453--5460.
\newblock Survey Track.

\bibitem[{Ferron et~al.(2023)Ferron, Shore, Mitra, and Agrawal}]{ferron-etal-2023-meep}
A.~Ferron, A.~Shore, E.~Mitra, and A.~Agrawal. 2023.
\newblock \href {https://doi.org/10.18653/v1/2023.findings-emnlp.137} {{MEEP}: Is this engaging? prompting large language models for dialogue evaluation in multilingual settings}.
\newblock In \emph{Findings of the Association for Computational Linguistics: EMNLP 2023}, pages 2078--2100, Singapore. Association for Computational Linguistics.

\bibitem[{Ferschke et~al.(2012)Ferschke, Gurevych, and Chebotar}]{ferschke-etal-2012-behind}
O.~Ferschke, I.~Gurevych, and Y.~Chebotar. 2012.
\newblock \href {https://aclanthology.org/E12-1079/} {Behind the article: Recognizing dialog acts in {W}ikipedia talk pages}.
\newblock In \emph{Proceedings of the 13th Conference of the {E}uropean Chapter of the Association for Computational Linguistics}, pages 777--786, Avignon, France.

\bibitem[{Fishkin et~al.(2018)Fishkin, Garg, Gelauff, Goel, Munagala, Sakshuwong, Siu, and Yandamuri}]{fishkin2018deliberative}
J.~Fishkin, N.~Garg, L.~Gelauff, A.~Goel, K.~Munagala, S.~Sakshuwong, A.~Siu, and S.~Yandamuri. 2018.
\newblock Deliberative democracy with the online deliberation platform.
\newblock In \emph{The 7th AAAI Conference on Human Computation and Crowdsourcing (HCOMP 2019). HCOMP}.

\bibitem[{Fournier-Tombs and MacKenzie(2021)}]{Fournier-Tombs21}
E.~Fournier-Tombs and M.~K. MacKenzie. 2021.
\newblock \href {https://doi.org/10.1177/20597991211010416} {Big data and democratic speech: Predicting deliberative quality using machine learning techniques}.
\newblock \emph{Methodological Innovations}, 14(2):20597991211010416.

\bibitem[{Friess(2018)}]{friess2018letting}
D.~M. Friess. 2018.
\newblock Letting the faculty deliberate: Analyzing online deliberation in academia using a comprehensive approach.
\newblock \emph{Journal of Information Technology \& Politics}, 15(2):155--177.

\bibitem[{Furniturewala and Jaidka(2024)}]{furniturewala-jaidka-2024-empaths}
S.~Furniturewala and K.~Jaidka. 2024.
\newblock \href {https://doi.org/10.18653/v1/2024.wassa-1.35} {Empaths at {WASSA} 2024 empathy and personality shared task: Turn-level empathy prediction using psychological indicators}.
\newblock In \emph{Proceedings of the 14th Workshop on Computational Approaches to Subjectivity, Sentiment, {\&} Social Media Analysis}, pages 404--411, Bangkok, Thailand. Association for Computational Linguistics.

\bibitem[{Gao et~al.(2025)Gao, Chen, Frermann, and Lau}]{gao-etal-2025-moderation}
R.~Gao, M.~Chen, L.~Frermann, and J.~H. Lau. 2025.
\newblock \href {https://doi.org/10.18653/v1/2025.findings-acl.106} {Moderation matters: Measuring conversational moderation impact in {E}nglish as a second language group discussion}.
\newblock In \emph{Findings of the Association for Computational Linguistics: ACL 2025}, pages 2070--2095, Vienna, Austria. Association for Computational Linguistics.

\bibitem[{Gelauff et~al.(2023)Gelauff, Nikolenko, Sakshuwong, Fishkin, Goel, Munagala, and Siu}]{gelauff_achieving_parity}
L.~Gelauff, L.~Nikolenko, S.~Sakshuwong, J.~Fishkin, A.~Goel, K.~Munagala, and A.~Siu. 2023.
\newblock \href {https://doi.org/10.4324/9781003215929-15} {\emph{Achieving parity with human moderators}}, pages 202--221.
\newblock Routledge.

\bibitem[{Gelula(1997)}]{Gelula1997}
M.~H. Gelula. 1997.
\newblock \href {https://doi.org/10.1016/s0090-3019(96)00342-4} {Clinical discussion sessions and small groups}.
\newblock \emph{Surgical Neurology}, 47(4):399--402.

\bibitem[{Gerber et~al.(2018)Gerber, Bächtiger, Shikano, Reber, and Rohr}]{gerber2018deliberative}
M.~Gerber, A.~Bächtiger, S.~Shikano, S.~Reber, and S.~Rohr. 2018.
\newblock Deliberative abilities and influence in a transnational deliberative poll (europolis).
\newblock \emph{British Journal of Political Science}, 48(4):1093--1118.

\bibitem[{Gimpel et~al.(2024)Gimpel, Lahmer, Wöhl et~al.}]{Gimpel2024}
H.~Gimpel, S.n Lahmer, M.~Wöhl, et~al. 2024.
\newblock \href {https://doi.org/10.1007/s10726-023-09856-8} {Digital facilitation of group work to gain predictable performance}.
\newblock \emph{Group Decision and Negotiation}, 33:113--145.

\bibitem[{Gligori'c et~al.(2024)Gligori'c, Zrnic, Lee, Candes, and Jurafsky}]{Gligoric2024CanUL}
K.~Gligori'c, T.~Zrnic, C.~Lee, E.~J. Candes, and D.~Jurafsky. 2024.
\newblock \href {https://api.semanticscholar.org/CorpusID:271962879} {Can unconfident llm annotations be used for confident conclusions?}
\newblock \emph{ArXiv}, abs/2408.15204.

\bibitem[{Goñi(2024)}]{Goni2024}
J.~I. Goñi. 2024.
\newblock \href {https://doi.org/10.1007/s11024-024-09551-1} {What is “dialogue” in public engagement with science and technology? bridging sts and deliberative democracy}.
\newblock \emph{Minerva}.

\bibitem[{Graham(2008)}]{graham2008disagree}
P.~Graham. 2008.
\newblock \href {https://paulgraham.com/disagree.html} {How to disagree}.
\newblock Accessed: 2024-06-24.

\bibitem[{Graham and Witschge(2003)}]{graham2003}
T.~Graham and T~Witschge. 2003.
\newblock \href {https://doi.org/doi:10.1515/comm.2003.012} {In search of online deliberation: Towards a new method for examining the quality of online discussions}.
\newblock \emph{Communications}, 28(2):173--204.

\bibitem[{Grice(1975)}]{grice1975logic}
H.~P. Grice. 1975.
\newblock Logic and conversation.
\newblock \emph{Syntax and semantics}, 3.

\bibitem[{Habibi et~al.(2024)Habibi, Hovy, and Schwarz}]{habibi2024contentmoderatorsdilemmaremoval}
M.~Habibi, D.~Hovy, and C.~Schwarz. 2024.
\newblock \href {https://arxiv.org/abs/2412.16114} {The content moderator's dilemma: Removal of toxic content and distortions to online discourse}.
\newblock \emph{Preprint}, arXiv:2412.16114.

\bibitem[{Hessel and Lee(2019)}]{hessel-lee-2019-somethings}
J.~Hessel and L.~Lee. 2019.
\newblock \href {https://doi.org/10.18653/v1/N19-1166} {Something`s brewing! early prediction of controversy-causing posts from discussion features}.
\newblock In \emph{Proceedings of the 2019 Conference of the North {A}merican Chapter of the Association for Computational Linguistics: Human Language Technologies, Volume 1 (Long and Short Papers)}, pages 1648--1659, Minneapolis, Minnesota. Association for Computational Linguistics.

\bibitem[{Hewitt et~al.(2024)Hewitt, Ashokkumar, Ghezae, and Willer}]{hewitt2024predicting}
L.~Hewitt, A.~Ashokkumar, I.~Ghezae, and R.~Willer. 2024.
\newblock Predicting results of social science experiments using large language models.
\newblock Equal contribution, order randomized.

\bibitem[{Hoque and Carenini(2016)}]{10.1145/2856767.2856782}
E.~Hoque and G.~Carenini. 2016.
\newblock \href {https://doi.org/10.1145/2856767.2856782} {Multiconvis: A visual text analytics system for exploring a collection of online conversations}.
\newblock In \emph{Proceedings of the 21st International Conference on Intelligent User Interfaces}, IUI '16, page 96–107, New York, NY, USA.

\bibitem[{Hu et~al.(2023)Hu, Floyd, Jouravlev, Fedorenko, and Gibson}]{hu-etal-2023-fine}
J.~Hu, S.~Floyd, O.~Jouravlev, E.~Fedorenko, and E.~Gibson. 2023.
\newblock \href {https://doi.org/10.18653/v1/2023.acl-long.230} {A fine-grained comparison of pragmatic language understanding in humans and language models}.
\newblock In \emph{Proceedings of the 61st Annual Meeting of the Association for Computational Linguistics (Volume 1: Long Papers)}, pages 4194--4213, Toronto, Canada.

\bibitem[{Hua et~al.(2018)Hua, Danescu-Niculescu-Mizil, Taraborelli, Thain, Sorensen, and Dixon}]{hua2018wikiconvcorpuscompleteconversational}
Y.~Hua, C.~Danescu-Niculescu-Mizil, D.~Taraborelli, N.~Thain, J.~Sorensen, and L.~Dixon. 2018.
\newblock \href {https://doi.org/10.18653/v1/D18-1305} {{W}iki{C}onv: A corpus of the complete conversational history of a large online collaborative community}.
\newblock In \emph{Proceedings of the 2018 Conference on Empirical Methods in Natural Language Processing}, pages 2818--2823, Brussels, Belgium.

\bibitem[{Irani et~al.(2024)Irani, Faloutsos, and Esterling}]{irani2024argusense}
A.~Irani, M.~Faloutsos, and K.~Esterling. 2024.
\newblock Argusense: Argument-centric analysis of online discourse.
\newblock In \emph{Proceedings of the International AAAI Conference on Web and Social Media}, volume~18, pages 663--675.

\bibitem[{Jansen et~al.(2023)Jansen, Jung, and Salminen}]{jansen_2023}
B.~J. Jansen, S.~Jung, and J.~Salminen. 2023.
\newblock \href {https://doi.org/10.1016/j.nlp.2023.100020} {Employing large language models in survey research}.
\newblock \emph{Natural Language Processing Journal}, 4:100020.

\bibitem[{Jin et~al.(2024)Jin, Zhang, Meng, Wang, and Tan}]{Jin2024}
H.~Jin, Y.~Zhang, D.~Meng, J.~Wang, and J.~Tan. 2024.
\newblock \href {https://doi.org/10.48550/ARXIV.2403.02901} {A comprehensive survey on process-oriented automatic text summarization with exploration of llm-based methods}.
\newblock \emph{arXiv preprint}.

\bibitem[{Jwalapuram(2017)}]{jwalapuram-2017-evaluating}
P.~Jwalapuram. 2017.
\newblock \href {https://doi.org/10.26615/issn.1314-9156.2017_003} {Evaluating dialogs based on {G}rice`s maxims}.
\newblock In \emph{Proceedings of the Student Research Workshop Associated with {RANLP} 2017}, pages 17--24, Varna.

\bibitem[{Kaner et~al.(2007)Kaner, Lind, Toldi, Fisk, and Berger}]{kaner2007facilitators}
S.~Kaner, Le. Lind, C.~Toldi, S.~Fisk, and D.~Berger. 2007.
\newblock \emph{Facilitator's Guide to Participatory Decision-Making}.
\newblock John Wiley \& Sons/Jossey-Bass, San Francisco.

\bibitem[{Kang and Qian(2024)}]{kang-qian-2024-implanting}
H.~Kang and T.~Qian. 2024.
\newblock \href {https://doi.org/10.18653/v1/2024.findings-acl.56} {Implanting {LLM}{'}s knowledge via reading comprehension tree for toxicity detection}.
\newblock In \emph{Findings of the Association for Computational Linguistics ACL 2024}, pages 947--962, Bangkok, Thailand and virtual meeting.

\bibitem[{Karadzhov et~al.(2021)Karadzhov, Stafford, and Vlachos}]{karadzhov2023delidata}
G.~Karadzhov, T.~Stafford, and A.~Vlachos. 2021.
\newblock \href {https://api.semanticscholar.org/CorpusID:236975941} {Delidata: A dataset for deliberation in multi-party problem solving}.
\newblock \emph{Proceedings of the ACM on Human-Computer Interaction}, 7:1 -- 25.

\bibitem[{Khalid and Lee(2022)}]{khalid-lee-2022-explaining}
B.~Khalid and S.~Lee. 2022.
\newblock \href {https://doi.org/10.18653/v1/2022.naacl-main.430} {Explaining dialogue evaluation metrics using adversarial behavioral analysis}.
\newblock In \emph{Proceedings of the 2022 Conference of the North American Chapter of the Association for Computational Linguistics: Human Language Technologies}, pages 5871--5883, Seattle, United States. Association for Computational Linguistics.

\bibitem[{Kies(2022)}]{kie2022online}
R.~Kies. 2022.
\newblock Online deliberative matrix.
\newblock In \emph{Research Methods in Deliberative Democracy}, pages 148--162. Oxford University Press.

\bibitem[{Kim et~al.(2021)Kim, Eun, Seering, and Lee}]{kim_et_al_chatbot}
S.~Kim, J.~Eun, J.~Seering, and J.~Lee. 2021.
\newblock \href {https://doi.org/10.1145/3449161} {Moderator chatbot for deliberative discussion: Effects of discussion structure and discussant facilitation}.
\newblock \emph{Proc. ACM Hum.-Comput. Interact.}, 5(CSCW1).

\bibitem[{Kumar et~al.(2024)Kumar, AbuHashem, and Durumeric}]{Kumar_AbuHashem_Durumeric_2024}
D.~Kumar, Y.~A. AbuHashem, and Z.~Durumeric. 2024.
\newblock Watch your language: Investigating content moderation with large language models.
\newblock \emph{Proceedings of the International AAAI Conference on Web and Social Media}, 18(1):865--878.

\bibitem[{Lambert et~al.(2024)Lambert, Choi, and Chandrasekharan}]{Lambert2024}
.C~Lambert, F.~Choi, and E.~Chandrasekharan. 2024.
\newblock \href {https://doi.org/10.1145/3686929} {"positive reinforcement helps breed positive behavior": Moderator perspectives on encouraging desirable behavior}.
\newblock \emph{Proc. ACM Hum.-Comput. Interact.}, 8(CSCW2).

\bibitem[{Langevin et~al.(2021)Langevin, Lordon, Avrahami, Cowan, Hirsch, and Hsieh}]{10.1145/3411764.3445312}
R.~Langevin, R.~J Lordon, T.~Avrahami, B.~R. Cowan, T.~Hirsch, and G.~Hsieh. 2021.
\newblock \href {https://doi.org/10.1145/3411764.3445312} {Heuristic evaluation of conversational agents}.
\newblock In \emph{Proceedings of the 2021 CHI Conference on Human Factors in Computing Systems}, New York, NY, USA.

\bibitem[{Lawrence et~al.(2017)Lawrence, Park, Budzynska, Cardie, Konat, and Reed}]{10.1145/3032989}
J.~Lawrence, J.~Park, K.~Budzynska, C.~Cardie, B.~Konat, and C.~Reed. 2017.
\newblock \href {https://doi.org/10.1145/3032989} {Using argumentative structure to interpret debates in online deliberative democracy and erulemaking}.
\newblock \emph{ACM Trans. Internet Technol.}, 17(3).

\bibitem[{Lawrence and Reed(2020)}]{10.1162/coli_a_00364}
J.~Lawrence and C.~Reed. 2020.
\newblock \href {https://doi.org/10.1162/coli_a_00364} {Argument mining: A survey}.
\newblock \emph{Computational Linguistics}, 45(4):765--818.

\bibitem[{Li et~al.(2021)Li, Zhang, Fei, Feng, and Zhou}]{li-etal-2021-conversations}
Z.~Li, J.~Zhang, Z.~Fei, Y.~Feng, and J.~Zhou. 2021.
\newblock \href {https://doi.org/10.18653/v1/2021.acl-long.11} {Conversations are not flat: Modeling the dynamic information flow across dialogue utterances}.
\newblock In \emph{Proceedings of the 59th Annual Meeting of the Association for Computational Linguistics and the 11th International Joint Conference on Natural Language Processing (Volume 1: Long Papers)}, pages 128--138, Online.

\bibitem[{Lim et~al.(2011)Lim, Cheung, and Hew}]{cheung-et-al-2011}
S.C.R. Lim, W.~Cheung, and K.~Hew. 2011.
\newblock Critical thinking in asynchronous online discussion: An investigation of student facilitation techniques.
\newblock \emph{New Horizons in Education}, 59:52--65.

\bibitem[{Lipman(2003)}]{Lipman_2003}
M.~Lipman. 2003.
\newblock \emph{Thinking in Education}, 2 edition.
\newblock Cambridge University Press.

\bibitem[{Liu et~al.(2023)Liu, Ultes, Minker, and Maier}]{ye23}
Y.~Liu, S.~Ultes, W.~Minker, and W.~Maier. 2023.
\newblock \href {https://doi.org/10.1145/3571884.3597125} {Unified conversational models with system-initiated transitions between chit-chat and task-oriented dialogues}.
\newblock In \emph{Proceedings of the 5th International Conference on Conversational User Interfaces}, CUI '23, New York, NY, USA.

\bibitem[{Lo and McAvoy(2023)}]{Lo_McAvoy_2023}
J.~Lo and P.~McAvoy. 2023.
\newblock \emph{Debate and Deliberation in Democratic Education}, page 298–310.
\newblock Cambridge Handbooks in Education. Cambridge University Press.

\bibitem[{Lugini et~al.(2020)Lugini, Olshefski, Singh, Litman, and Godley}]{lugini-etal-2020-discussion}
L.~Lugini, C.~Olshefski, R.~Singh, D.~Litman, and A.~Godley. 2020.
\newblock \href {https://doi.org/10.18653/v1/2020.coling-demos.10} {Discussion tracker: Supporting teacher learning about students' collaborative argumentation in high school classrooms}.
\newblock In \emph{Proceedings of the 28th International Conference on Computational Linguistics: System Demonstrations}, pages 53--58, Barcelona, Spain (Online). International Committee on Computational Linguistics (ICCL).

\bibitem[{Macagno et~al.(2022)Macagno, Rapanta, Mayweg-Paus, and Garcia-Milà}]{MACAGNO2022116}
F.~Macagno, C.~Rapanta, E.~Mayweg-Paus, and M.~Garcia-Milà. 2022.
\newblock \href {https://doi.org/10.1016/j.pragma.2022.02.011} {Coding empathy in dialogue}.
\newblock \emph{Journal of Pragmatics}, 192:116--132.

\bibitem[{Mansour(2024)}]{Mansour2024}
N.~Mansour. 2024.
\newblock \href {https://doi.org/10.1007/s10639-024-12473-w} {Students’ and facilitators’ experiences with synchronous and asynchronous online dialogic discussions and e-facilitation in understanding the nature of science}.
\newblock \emph{Education and Information Technologies}, 29:15965--15997.

\bibitem[{Martinenghi et~al.(2024)Martinenghi, Donabauer, Amenta, Bursic, Giudici, Kruschwitz, Garzotto, and Ognibene}]{martinenghi-etal-2024-llms}
A.~Martinenghi, G.~Donabauer, S.~Amenta, S.~Bursic, M.~Giudici, U.~Kruschwitz, F.~Garzotto, and D.~Ognibene. 2024.
\newblock \href {https://aclanthology.org/2024.games-1.12/} {{LLM}s of catan: Exploring pragmatic capabilities of generative chatbots through prediction and classification of dialogue acts in boardgames' multi-party dialogues}.
\newblock In \emph{Proceedings of the 10th Workshop on Games and Natural Language Processing @ LREC-COLING 2024}, pages 107--118, Torino, Italia.

\bibitem[{Mathew et~al.(2019)Mathew, Dutt, Goyal, and Mukherjee}]{binny2019}
B.~Mathew, R.~Dutt, P.~Goyal, and A.~Mukherjee. 2019.
\newblock Spread of hate speech in online social media.
\newblock In \emph{Proceedings of the 10th ACM Conference on Web Science}, page 173–182, New York, NY, USA.

\bibitem[{Mendonca et~al.(2024)Mendonca, Trancoso, and Lavie}]{mendonca-etal-2024-ecoh}
J.~Mendonca, I.~Trancoso, and A.~Lavie. 2024.
\newblock \href {https://doi.org/10.18653/v1/2024.sigdial-1.44} {{EC}oh: Turn-level coherence evaluation for multilingual dialogues}.
\newblock In \emph{Proceedings of the 25th Annual Meeting of the Special Interest Group on Discourse and Dialogue}, pages 516--532, Kyoto, Japan.

\bibitem[{Mirzakhmedova et~al.(2024)Mirzakhmedova, Gohsen, Chang, and Stein}]{mirzakhmedova2024large}
N.~Mirzakhmedova, M.~Gohsen, C.~H. Chang, and B.~Stein. 2024.
\newblock Are large language models reliable argument quality annotators?
\newblock In \emph{Conference on Advances in Robust Argumentation Machines}, pages 129--146. Springer.

\bibitem[{{MIT Center for Constructive Communication}(2024)}]{dimitra-guide}
{MIT Center for Constructive Communication}. 2024.
\newblock Unpublished training materials developed by the mit center for constructive communication.
\newblock Guide given to human facilitators.

\bibitem[{Molina and Sundar(2022)}]{Molina2022}
M.D. Molina and S.S. Sundar. 2022.
\newblock {When AI moderates online content: effects of human collaboration and interactive transparency on user trust}.
\newblock \emph{Journal of Computer-Mediated Communication}, 27(4).

\bibitem[{Nam et~al.(2023)Nam, Chung, and Hong}]{doi:10.1089/cyber.2022.0356}
Y.~Nam, H.~Chung, and U.~Hong. 2023.
\newblock \href {https://doi.org/10.1089/cyber.2022.0356} {Language artificial intelligences' communicative performance quantified through the gricean conversation theory}.
\newblock \emph{Cyberpsychology, Behavior, and Social Networking}, 26(12):919--923.
\newblock PMID: 37976199.

\bibitem[{Neumann et~al.(2025)Neumann, De-Arteaga, and Fazelpour}]{neumann_2025}
T.~Neumann, M.~De-Arteaga, and S.~Fazelpour. 2025.
\newblock \href {https://arxiv.org/abs/2504.08954} {Should you use llms to simulate opinions? quality checks for early-stage deliberation}.
\newblock \emph{Preprint}, arXiv:2504.08954.

\bibitem[{Ngai et~al.(2021)Ngai, Lee, Luo, Chan, and Liang}]{NGAI2021101098}
E.W.T. Ngai, M.C.M. Lee, M.~Luo, P.S.L. Chan, and T.~Liang. 2021.
\newblock \href {https://doi.org/10.1016/j.elerap.2021.101098} {An intelligent knowledge-based chatbot for customer service}.
\newblock \emph{Electronic Commerce Research and Applications}, 50:101098.

\bibitem[{Niculae and Danescu-Niculescu-Mizil(2016)}]{niculae-danescu-niculescu-mizil-2016-conversational}
V.~Niculae and C.~Danescu-Niculescu-Mizil. 2016.
\newblock \href {https://doi.org/10.18653/v1/N16-1070} {Conversational markers of constructive discussions}.
\newblock In \emph{Proceedings of the 2016 Conference of the North {A}merican Chapter of the Association for Computational Linguistics: Human Language Technologies}, pages 568--578, San Diego, California.

\bibitem[{Park et~al.(2012)Park, Klingel, Cardie, Newhart, Farina, and Vallb\'{e}}]{park_et_al_2012_facilitation}
J.~Park, S.~Klingel, C.~Cardie, M.~Newhart, C.~Farina, and J.J. Vallb\'{e}. 2012.
\newblock \href {https://doi.org/10.1145/2307729.2307757} {Facilitative moderation for online participation in erulemaking}.
\newblock In \emph{Proceedings of the 13th Annual International Conference on Digital Government Research}, page 173–182, New York, NY, USA.

\bibitem[{Park et~al.(2023)Park, O'Brien, Cai, Morris, Liang, and Bernstein}]{Park2023GenerativeAI}
J.S. Park, J.C. O'Brien, C.J. Cai, M.R. Morris, P.~Liang, and M.S. Bernstein. 2023.
\newblock \href {https://api.semanticscholar.org/CorpusID:258040990} {Generative agents: Interactive simulacra of human behavior}.
\newblock \emph{Proceedings of the 36th Annual ACM Symposium on User Interface Software and Technology}.

\bibitem[{Park et~al.(2022)Park, Popowski, Cai, Morris, Liang, and Bernstein}]{park2022socialsimulacracreatingpopulated}
J.S. Park, L.~Popowski, C.J. Cai, M.R. Morris, P.~Liang, and M.S. Bernstein. 2022.
\newblock \href {https://doi.org/10.1145/3526113.3545616} {Social simulacra: Creating populated prototypes for social computing systems}.
\newblock In \emph{Proceedings of the 35th Annual ACM Symposium on User Interface Software and Technology}, UIST '22, New York, NY, USA.

\bibitem[{Pradel et~al.(2024)Pradel, Zilinsky, Kosmidis, and Theocharis}]{PRADEL_ZILINSKY_KOSMIDIS_THEOCHARIS_2024}
F.~Pradel, J.~Zilinsky, S.~Kosmidis, and Y.~Theocharis. 2024.
\newblock \href {https://doi.org/10.1017/S000305542300134X} {Toxic speech and limited demand for content moderation on social media}.
\newblock \emph{American Political Science Review}, 118(4):1895–1912.

\bibitem[{Raj~Prabhu et~al.(2021)Raj~Prabhu, Raman, and Hung}]{prabhu21}
N.~Raj~Prabhu, C.~Raman, and H.~Hung. 2021.
\newblock \href {https://doi.org/10.1145/3395035.3425966} {Defining and quantifying conversation quality in spontaneous interactions}.
\newblock In \emph{Companion Publication of the 2020 International Conference on Multimodal Interaction}, ICMI '20 Companion, page 196–205, New York, NY, USA.

\bibitem[{Rescala et~al.(2024)Rescala, Ribeiro, Hu, and West}]{rescala-etal-2024-language}
P.~Rescala, M.H. Ribeiro, T.~Hu, and R.~West. 2024.
\newblock \href {https://doi.org/10.18653/v1/2024.findings-emnlp.515} {Can language models recognize convincing arguments?}
\newblock In \emph{Findings of the Association for Computational Linguistics: EMNLP 2024}, pages 8826--8837, Miami, Florida, USA.

\bibitem[{Rheingold(2000)}]{rheingold00}
H.~Rheingold. 2000.
\newblock \emph{{The Virtual Community: Homesteading on the Electronic Frontier}}.
\newblock The MIT Press.

\bibitem[{Rose-Redwood et~al.(2018)Rose-Redwood, Kitchin, Rickards, Rossi, Datta, and Crampton}]{Redwood18}
R.~Rose-Redwood, R.~Kitchin, L.~Rickards, U.~Rossi, A.~Datta, and J.~Crampton. 2018.
\newblock \href {https://doi.org/10.1177/2043820618780566} {The possibilities and limits to dialogue}.
\newblock \emph{Dialogues in Human Geography}, 8(2):109--123.

\bibitem[{Rossi et~al.(2024)Rossi, Harrison, and Shklovski}]{rossi_2024}
L.~Rossi, K.~Harrison, and I.~Shklovski. 2024.
\newblock \href {https://doi.org/10.6092/issn.1971-8853/19576} {The problems of llm-generated data in social science research}.
\newblock \emph{Sociologica}, 18(2):145–168.

\bibitem[{Ruis et~al.(2023)Ruis, Khan, Biderman, Hooker, Rockt\"{a}schel, and Grefenstette}]{10.5555/3666122.3667035}
L.~Ruis, A.~Khan, S.~Biderman, S.~Hooker, T.~Rockt\"{a}schel, and E.~Grefenstette. 2023.
\newblock The goldilocks of pragmatic understanding: fine-tuning strategy matters for implicature resolution by llms.
\newblock In \emph{Proceedings of the 37th International Conference on Neural Information Processing Systems}, NIPS '23, Red Hook, NY, USA.

\bibitem[{Russmann and Lane(2016)}]{Russman16}
U.~Russmann and A.~Lane. 2016.
\newblock \href {https://ijoc.org/index.php/ijoc/article/view/6086} {Discussion. dialogue, and discourse| doing the talk: Discussion, dialogue, and discourse in action — introduction}.
\newblock \emph{International Journal of Communication}, 10.

\bibitem[{Saeidi et~al.(2021)Saeidi, Yazdani, and Vlachos}]{saeidi-etal-2021-cross}
M.~Saeidi, M.~Yazdani, and A.~Vlachos. 2021.
\newblock Cross-policy compliance detection via question answering.
\newblock In \emph{Proceedings of the 2021 Conference on Empirical Methods in Natural Language Processing}, pages 8622--8632, Online and Punta Cana, Dominican Republic.

\bibitem[{Sak et~al.(2014)Sak, Senior, and Beaufays}]{Sak2014LongSM}
H.~Sak, A.~W. Senior, and F.~Beaufays. 2014.
\newblock \href {https://api.semanticscholar.org/CorpusID:16904319} {Long short-term memory based recurrent neural network architectures for large vocabulary speech recognition}.
\newblock \emph{ArXiv}, abs/1402.1128.

\bibitem[{Sap et~al.(2020)Sap, Gabriel, Qin, Jurafsky, Smith, and Choi}]{sap-etal-2020-social}
M.~Sap, S.~Gabriel, L.~Qin, D.~Jurafsky, N.~A. Smith, and Y.~Choi. 2020.
\newblock \href {https://doi.org/10.18653/v1/2020.acl-main.486} {Social bias frames: Reasoning about social and power implications of language}.
\newblock In \emph{Proceedings of the 58th Annual Meeting of the Association for Computational Linguistics}, pages 5477--5490, Online.

\bibitem[{Schaffner et~al.(2024)Schaffner, Bhagoji, Cheng, Mei, Shen, Wang, Chetty, Feamster, Lakier, and Tan}]{schaffner_community_guidelines}
B.~Schaffner, A.~N. Bhagoji, S.~Cheng, J.~Mei, J.L. Shen, G.~Wang, M.~Chetty, N.~Feamster, G.~Lakier, and C.~Tan. 2024.
\newblock \href {https://doi.org/10.1145/3613904.3642333} {"{C}ommunity guidelines make this the best party on the internet": An in-depth study of online platforms' content moderation policies}.
\newblock In \emph{Proceedings of the 2024 CHI Conference on Human Factors in Computing Systems}, CHI '24, New York, NY, USA.

\bibitem[{Schluger et~al.(2022)Schluger, Chang, Danescu-Niculescu-Mizil, and Levy}]{proactive_moderation}
C.~Schluger, J.P. Chang, C.~Danescu-Niculescu-Mizil, and K.~Levy. 2022.
\newblock \href {https://doi.org/10.1145/3555095} {Proactive moderation of online discussions: Existing practices and the potential for algorithmic support}.
\newblock \emph{Proc. ACM Hum.-Comput. Interact.}, 6(CSCW2).

\bibitem[{Schroeder et~al.(2024)Schroeder, Roy, and Kabbara}]{schroeder-etal-2024-fora}
H.~Schroeder, D.~Roy, and J.~Kabbara. 2024.
\newblock Fora: A corpus and framework for the study of facilitated dialogue.
\newblock In \emph{Proceedings of the 62nd Annual Meeting of the Association for Computational Linguistics}, pages 13985--14001, Bangkok, Thailand.

\bibitem[{Searle(1969)}]{Searle_1969}
J.~R. Searle. 1969.
\newblock \emph{Speech Acts: An Essay in the Philosophy of Language}.
\newblock Cambridge University Press.

\bibitem[{See et~al.(2019)See, Roller, Kiela, and Weston}]{see-etal-2019-makes}
A.~See, S.~Roller, D.~Kiela, and J.~Weston. 2019.
\newblock \href {https://doi.org/10.18653/v1/N19-1170} {What makes a good conversation? how controllable attributes affect human judgments}.
\newblock In \emph{Proceedings of the 2019 Conference of the North {A}merican Chapter of the Association for Computational Linguistics: Human Language Technologies, Volume 1 (Long and Short Papers)}, pages 1702--1723, Minneapolis, Minnesota. Association for Computational Linguistics.

\bibitem[{Seering(2020)}]{seering_self_moderation}
J.~Seering. 2020.
\newblock \href {https://doi.org/10.1145/3415178} {Reconsidering self-moderation: the role of research in supporting community-based models for online content moderation}.
\newblock \emph{Proc. ACM Hum.-Comput. Interact.}, 4(CSCW2).

\bibitem[{Shahid et~al.(2024)Shahid, Dittgen, Naaman, and Vashistha}]{shahid2024examining}
F.~Shahid, M.~Dittgen, M.~Naaman, and A.~Vashistha. 2024.
\newblock Examining human-{AI} collaboration for co-writing constructive comments online.
\newblock \emph{arXiv preprint arXiv:2411.03295}.

\bibitem[{Shi et~al.(2024)Shi, Liu, and Song}]{shi-2024-hatespeech}
X.~Shi, J.~Liu, and Y.~Song. 2024.
\newblock {BERT} and {LLM}-based multivariate hate speech detection on twitter: Comparative analysis and superior performance.
\newblock In \emph{Artificial Intelligence and Machine Learning}, pages 85--97, Singapore. Springer Nature Singapore.

\bibitem[{Sjølie et~al.(2021)Sjølie, Strømme, and Boks-Vlemmix}]{SJOLIE2021103477}
E.~Sjølie, A.~Strømme, and J.~Boks-Vlemmix. 2021.
\newblock \href {https://doi.org/10.1016/j.tate.2021.103477} {Team-skills training and real-time facilitation as a means for developing student teachers’ learning of collaboration}.
\newblock \emph{Teaching and Teacher Education}, 107:103477.

\bibitem[{Small et~al.(2023)Small, Vendrov, Durmus, Homaei, Barry, Cornebise, Suzman, Ganguli, and Megill}]{small-polis-llm}
C.T. Small, I.~Vendrov, E.~Durmus, H.~Homaei, E.~Barry, J.~Cornebise, T.~Suzman, D.~Ganguli, and C.~Megill. 2023.
\newblock Opportunities and risks of {LLMs} for scalable deliberation with {P}olis.
\newblock \emph{ArXiv}, abs/2306.11932.

\bibitem[{Smith et~al.(2022)Smith, Hsu, Qian, Roller, Boureau, and Weston}]{smith-etal-2022-human}
E.~Smith, O.~Hsu, R.~Qian, S.~Roller, Y-L. Boureau, and J.~Weston. 2022.
\newblock \href {https://doi.org/10.18653/v1/2022.nlp4convai-9} {Human evaluation of conversations is an open problem: Comparing the sensitivity of various methods for evaluating dialogue agents}.
\newblock In \emph{Proceedings of the 4th Workshop on NLP for Conversational AI}, pages 77--97, Dublin, Ireland. Association for Computational Linguistics.

\bibitem[{Sravanthi et~al.(2024)Sravanthi, Doshi, Tankala, Murthy, Dabre, and Bhattacharyya}]{sravanthi-etal-2024-pub}
S.~Sravanthi, M.~Doshi, P.~Tankala, R.~Murthy, R.~Dabre, and P.~Bhattacharyya. 2024.
\newblock \href {https://doi.org/10.18653/v1/2024.findings-acl.719} {{PUB}: A pragmatics understanding benchmark for assessing {LLM}s' pragmatics capabilities}.
\newblock In \emph{Findings of the Association for Computational Linguistics: ACL 2024}, pages 12075--12097, Bangkok, Thailand.

\bibitem[{Steenbergen et~al.(2003)Steenbergen, Bächtiger, Spörndli, and Steiner}]{steenbergen2003measuring}
M.~Steenbergen, A.~Bächtiger, M.~Spörndli, and J.~Steiner. 2003.
\newblock \href {https://doi.org/10.1057/palgrave.cep.6110002} {Measuring political deliberation: A discourse quality index}.
\newblock \emph{Comparative European Politics}, 1:21--48.

\bibitem[{Stolcke et~al.(2000)Stolcke, Ries, Coccaro, Shriberg, Bates, Jurafsky, Taylor, Martin, Van Ess-Dykema, and Meteer}]{stolcke-etal-2000-dialogue}
A.~Stolcke, K.~Ries, N.~Coccaro, E.~Shriberg, R.~Bates, D.~Jurafsky, P.~Taylor, R.~Martin, C.~Van Ess-Dykema, and M.~Meteer. 2000.
\newblock \href {https://aclanthology.org/J00-3003/} {Dialogue act modeling for automatic tagging and recognition of conversational speech}.
\newblock \emph{Computational Linguistics}, 26(3):339--374.

\bibitem[{Stromer-Galley(2007)}]{Stromer-Galley2007}
J.~Stromer-Galley. 2007.
\newblock Measuring deliberation’s content: A coding scheme.
\newblock \emph{Journal of Deliberative Democracy}, 3(1):25--44.

\bibitem[{Sun et~al.(2021)Sun, Moon, Crook, Roller, Silvert, Liu, Wang, Liu, Cho, and Cardie}]{sun-etal-2021-adding}
K.~Sun, S.~Moon, P.~Crook, S.~Roller, B.~Silvert, B.~Liu, Z.~Wang, H.~Liu, E.~Cho, and C.~Cardie. 2021.
\newblock \href {https://doi.org/10.18653/v1/2021.naacl-main.124} {Adding chit-chat to enhance task-oriented dialogues}.
\newblock In \emph{Proceedings of the 2021 Conference of the North American Chapter of the Association for Computational Linguistics: Human Language Technologies}, pages 1570--1583, Online. Association for Computational Linguistics.

\bibitem[{Tan et~al.(2016)Tan, Niculae, Danescu-Niculescu-Mizil, and Lee}]{10.1145/2872427.2883081}
C.~Tan, V.~Niculae, C.~Danescu-Niculescu-Mizil, and L.~Lee. 2016.
\newblock \href {https://doi.org/10.1145/2872427.2883081} {Winning arguments: Interaction dynamics and persuasion strategies in good-faith online discussions}.
\newblock In \emph{Proceedings of the 25th International Conference on World Wide Web}, WWW '16, page 613–624, Republic and Canton of Geneva, CHE.

\bibitem[{Taubenfeld et~al.(2024)Taubenfeld, Dover, Reichart, and Goldstein}]{Taubenfeld2024SystematicBI}
A.~Taubenfeld, Y.~Dover, R.~Reichart, and A.~Goldstein. 2024.
\newblock \href {https://api.semanticscholar.org/CorpusID:267499945} {Systematic biases in llm simulations of debates}.
\newblock \emph{ArXiv}, abs/2402.04049.

\bibitem[{Tessler et~al.(2024)Tessler, Bakker, Jarrett, Sheahan, Chadwick, Koster, Evans, Campbell-Gillingham, T.Collins, Parkes, Botvinick, and Summerfield}]{Tessler24}
M.H. Tessler, M.A. Bakker, D.~Jarrett, H.~Sheahan, M.J. Chadwick, R.~Koster, G.~Evans, L.~Campbell-Gillingham, T.Collins, D.C. Parkes, M.~Botvinick, and C.~Summerfield. 2024.
\newblock {AI} can help humans find common ground in democratic deliberation.
\newblock \emph{Science}, 386(6719).

\bibitem[{{The Commons}(2025)}]{thecommons2025}
{The Commons}. 2025.
\newblock \href {https://howtobuildup.org/programs/digital-conflict/the-commons-project/} {The commons project}.
\newblock Accessed: 2025-01-27.

\bibitem[{Trenel(2009)}]{trenel2009facilitation}
M.~Trenel. 2009.
\newblock Facilitation and inclusive deliberation.
\newblock In \emph{Online Deliberation: Design, Research, and Practice}, pages 253--257. CSLI Publications/University of Chicago Press.

\bibitem[{Trujillo and Cresci(2022)}]{make_reddit_great}
A.~Trujillo and S.~Cresci. 2022.
\newblock \href {https://api.semanticscholar.org/CorpusID:246016404} {Make reddit great again: Assessing community effects of moderation interventions on r/the\_donald}.
\newblock \emph{Proceedings of the ACM on Human-Computer Interaction}, 6:1 -- 28.

\bibitem[{Tsai et~al.(2024)Tsai, Pentland, Braley, Chen, Enr{\' i}quez, and Reuel}]{Tsai2024Generative}
L.~L. Tsai, A.~Pentland, A.~Braley, N.~Chen, J.~R. Enr{\' i}quez, and A.~Reuel. 2024.
\newblock Generative {AI} for {Pro}-{Democracy} {Platforms}.
\newblock \emph{An MIT Exploration of Generative AI}.
\newblock Https://mit-genai.pubpub.org/pub/mn45hexw.

\bibitem[{Tucker et~al.(2018)Tucker, Guess, Barberá, Vaccari, Siegel, Sanovich, Stukal, and Nyhan}]{tucker2018social}
J.A. Tucker, A.~Guess, P.~Barberá, C.~Vaccari, A.~Siegel, S.~Sanovich, D.~Stukal, and B.~Nyhan. 2018.
\newblock Social media, political polarization, and political disinformation: A review of the scientific literature.
\newblock \emph{SSRN Electronic Journal}.

\bibitem[{Ulmer et~al.(2024)Ulmer, Mansimov, Lin, Sun, Gao, and Zhang}]{ulmer2024bootstrappingllmbasedtaskorienteddialogue}
D.~Ulmer, E.~Mansimov, K.~Lin, L.~Sun, X.~Gao, and Y.~Zhang. 2024.
\newblock \href {https://doi.org/10.18653/v1/2024.findings-acl.566} {Bootstrapping {LLM}-based task-oriented dialogue agents via self-talk}.
\newblock In \emph{Findings of the Association for Computational Linguistics: ACL 2024}, pages 9500--9522, Bangkok, Thailand.

\bibitem[{Vecchi et~al.(2021)Vecchi, Falk, Jundi, and Lapesa}]{vecchi-etal-2021-towards}
E.M. Vecchi, N.~Falk, I.~Jundi, and G.~Lapesa. 2021.
\newblock Towards argument mining for social good: A survey.
\newblock In \emph{Proceedings of the 59th Annual Meeting of {ACL} and 11th International Joint Conference on {NLP}}, pages 1338--1352, Online.

\bibitem[{Veglis(2014)}]{veglis_moderation}
A.~Veglis. 2014.
\newblock Moderation techniques for social media content.
\newblock In \emph{Social Computing and Social Media}, pages 137--148, Cham. Springer International Publishing.

\bibitem[{Verkuyl et~al.(2024)Verkuyl, Violato, Southam, Lavoie-Tremblay, Goldsworthy, MacEachern, and Atack}]{Verkuyl2024}
M.~Verkuyl, E.~Violato, T.~Southam, M.~Lavoie-Tremblay, S.~Goldsworthy, D.~MacEachern, and L.~Atack. 2024.
\newblock \href {https://doi.org/10.1186/s41077-024-00323-1} {Facilitators' experiences with virtual simulation and their impact on learning}.
\newblock \emph{Advances in Simulation}, 9(1):54.

\bibitem[{Wachsmuth et~al.(2024)Wachsmuth, Lapesa, Cabrio, Lauscher, Park, Vecchi, Villata, and Ziegenbein}]{wachsmuth-etal-2024-argument}
H.~Wachsmuth, G.~Lapesa, E.~Cabrio, A.~Lauscher, J.~Park, E.M. Vecchi, S.~Villata, and T.~Ziegenbein. 2024.
\newblock Argument quality assessment in the age of instruction-following large language models.
\newblock In \emph{Proceedings of the 2024 Joint International Conference on Computational Linguistics, Language Resources and Evaluation}, pages 1519--1538, Torino, Italia.

\bibitem[{Wachsmuth et~al.(2017)Wachsmuth, Naderi, Hou, Bilu, Prabhakaran, Thijm, Hirst, and Stein}]{wachsmuth2017computational}
H.~Wachsmuth, N.~Naderi, Y.~Hou, Y.~Bilu, V.~Prabhakaran, T.A. Thijm, G.~Hirst, and B.~Stein. 2017.
\newblock Computational argumentation quality assessment in natural language.
\newblock In \emph{Proceedings of the 15th Conference of the European Chapter of the Association for Computational Linguistics: Volume 1, Long Papers}, pages 176--187.

\bibitem[{Walker et~al.(2012)Walker, Tree, Anand, Abbott, and King}]{walker-etal-2012-corpus}
M.~Walker, J.~F. Tree, P.~Anand, R.~Abbott, and J.~King. 2012.
\newblock A corpus for research on deliberation and debate.
\newblock In \emph{Proceedings of the Eighth International Conference on Language Resources and Evaluation ({LREC}'12)}, pages 812--817, Istanbul, Turkey.

\bibitem[{Wang and Chang(2022)}]{Wang2022ToxicityDW}
Y.~Wang and Y.~T. Chang. 2022.
\newblock \href {https://api.semanticscholar.org/CorpusID:249062985} {Toxicity detection with generative prompt-based inference}.
\newblock \emph{ArXiv}, abs/2205.12390.

\bibitem[{Wang et~al.(2023)Wang, Chen, He, and Sun}]{wang2023contextual}
Y.~Wang, X.~Chen, B.~He, and L.~Sun. 2023.
\newblock Contextual interaction for argument post quality assessment.
\newblock In \emph{Proceedings of the 2023 Conference on Empirical Methods in Natural Language Processing}, pages 10420--10432.

\bibitem[{Warner et~al.(2025)Warner, Strohmayer, Higgs, and Coventry}]{WARNER2025103468}
M.~Warner, A.~Strohmayer, M.~Higgs, and L.~Coventry. 2025.
\newblock \href {https://doi.org/10.1016/j.ijhcs.2025.103468} {A critical reflection on the use of toxicity detection algorithms in proactive content moderation systems}.
\newblock \emph{International Journal of Human-Computer Studies}, 198:103468.

\bibitem[{White et~al.(2024)White, Hunter, and Greaves}]{dimitra-book}
K.~White, N.~Hunter, and K.~Greaves. 2024.
\newblock \emph{facilitating deliberation - a practical guide}.
\newblock Mosaic Lab.

\bibitem[{Wilson et~al.(1984)Wilson, Wiemann, and Zimmerman}]{doi:10.1177/0261927X8400300301}
T.~P. Wilson, J.~M. Wiemann, and D.~H. Zimmerman. 1984.
\newblock \href {https://doi.org/10.1177/0261927X8400300301} {Models of turn taking in conversational interaction}.
\newblock \emph{Journal of Language and Social Psychology}, 3(3):159--183.

\bibitem[{Xu and Jiang(2024)}]{xu-jiang-2024-multi}
Z.~Xu and J.~Jiang. 2024.
\newblock \href {https://doi.org/10.18653/v1/2024.findings-emnlp.113} {Multi-dimensional evaluation of empathetic dialogue responses}.
\newblock In \emph{Findings of the Association for Computational Linguistics: EMNLP 2024}, pages 2066--2087, Miami, Florida, USA. Association for Computational Linguistics.

\bibitem[{Yeh et~al.(2021)Yeh, Eskenazi, and Mehri}]{yeh-etal-2021-comprehensive}
Y.~Yeh, M.~Eskenazi, and S.~Mehri. 2021.
\newblock \href {https://doi.org/10.18653/v1/2021.eancs-1.3} {A comprehensive assessment of dialog evaluation metrics}.
\newblock In \emph{The First Workshop on Evaluations and Assessments of Neural Conversation Systems}, pages 15--33, Online. Association for Computational Linguistics.

\bibitem[{Yu et~al.(2024)Yu, Li, Su, and Fuoli}]{yu2024}
D.~Yu, L.~Li, H.~Su, and M.~Fuoli. 2024.
\newblock \href {https://doi.org/10.1075/ijcl.23087.yu} {Assessing the potential of llm-assisted annotation for corpus-based pragmatics and discourse analysis}.
\newblock \emph{International Journal of Corpus Linguistics}, 29(4):534--561.

\bibitem[{Zhang et~al.(2017)Zhang, Culbertson, and Paritosh}]{Zhang_Culbertson_Paritosh_2017}
A.~Zhang, B.~Culbertson, and P.~Paritosh. 2017.
\newblock \href {https://doi.org/10.1609/icwsm.v11i1.14886} {Characterizing online discussion using coarse discourse sequences}.
\newblock \emph{Proceedings of the International AAAI Conference on Web and Social Media}, 11(1):357--366.

\bibitem[{Zhang et~al.(2023)Zhang, D{'}Haro, Tang, Shi, Tang, and Li}]{zhang-etal-2023-xdial}
C.~Zhang, L.~D{'}Haro, C.~Tang, K.~Shi, G.~Tang, and H.~Li. 2023.
\newblock \href {https://doi.org/10.18653/v1/2023.findings-emnlp.371} {x{D}ial-eval: A multilingual open-domain dialogue evaluation benchmark}.
\newblock In \emph{Findings of the Association for Computational Linguistics: EMNLP 2023}, pages 5579--5601, Singapore.

\bibitem[{Zhang et~al.(2024)Zhang, D'Haro, Chen, Zhang, and Li}]{zhang2024comprehensive}
C.~Zhang, L.~F. D'Haro, Y.~Chen, M.~Zhang, and H.~Li. 2024.
\newblock A comprehensive analysis of the effectiveness of large language models as automatic dialogue evaluators.
\newblock In \emph{Proceedings of the AAAI Conference on Artificial Intelligence}, volume~38, pages 19515--19524.

\bibitem[{Zhang et~al.(2018)Zhang, Chang, Danescu-Niculescu-Mizil, Dixon, Hua, Taraborelli, and Thain}]{zhang-etal-2018-conversations}
J.~Zhang, J.~Chang, C.~Danescu-Niculescu-Mizil, L.~Dixon, Y.~Hua, D.~Taraborelli, and N.~Thain. 2018.
\newblock \href {https://doi.org/10.18653/v1/P18-1125} {Conversations gone awry: Detecting early signs of conversational failure}.
\newblock In \emph{Proceedings of the 56th Annual Meeting of the Association for Computational Linguistics (Volume 1: Long Papers)}, pages 1350--1361, Melbourne, Australia.

\bibitem[{Zhou et~al.(2024)Zhou, Farag, and Vlachos}]{zhou-etal-2024-llm-feature}
L.~Zhou, Y.~Farag, and A.~Vlachos. 2024.
\newblock \href {https://doi.org/10.18653/v1/2024.emnlp-main.308} {An {LLM} feature-based framework for dialogue constructiveness assessment}.
\newblock In \emph{Proceedings of the 2024 Conference on Empirical Methods in Natural Language Processing}, pages 5389--5409, Miami, Florida, USA. Association for Computational Linguistics.

\bibitem[{Ziems et~al.(2024)Ziems, Held, Shaikh, Chen, Zhang, and Yang}]{ziems2024can}
C.~Ziems, W.~Held, O.~Shaikh, J.~Chen, Z.~Zhang, and D.~Yang. 2024.
\newblock \href {https://doi.org/10.1162/coli_a_00502} {Can large language models transform computational social science?}
\newblock \emph{Computational Linguistics}, 50(1):237--291.

\end{thebibliography}

\appendix

\label{sec:appendix}

\section{Acronyms}
\label{sec:appendix:acronyms}

\begin{acronym}[WWW] 
	\acro{NLP}{Natural Language Processing}
        \acro{ML}{Machine Learning}
	\acro{LLM}[LLM]{Large Language Model}
        \acro{AM}{Argument Mining}
	\acro{ML}{Machine Learning}
	\acro{IR}{Information Retrieval}
        \acro{AQ}{Argument Quality}
        \acro{ESL}{English as a Second Language}
\end{acronym}

\section {Keywords for Literature Query}
\label{sec:appendix:keywords}







\begin{table}[!h]
\centering
\small
\begin{tabular}{l}
\toprule
\textbf{Keyword Selection} \\ \midrule
online discussions, deliberation, dialogue, \\
discussion evaluation, discussion metrics,  \\
dialogue, deliberation, NLP, AI, discussion quality,\\
argument mining, survey, LLM, conversation, \\
moderation, facilitation, communication, democracy \\ 
AI dialogue systems, group dynamics \\
\bottomrule
\end{tabular}
\caption{Keywords for search engine queries}
\label{tab:keywords}
\end{table}

\section{Terminology Background}
\label{sec:appendix:terminology}
Here, we explain our 
reasoning for choosing and disambiguating certain terms (see \S\ref{sec:terminology}).
The definitions of the terms can be found in Table~\ref{tab:terminology}.

\paragraph{Facilitation vs. Moderation} “Moderation”, as a term, is more common in Computer Science and \ac{NLP}, while facilitation is prevalent in Social Sciences \cite{vecchi-etal-2021-towards, kaner2007facilitators, trenel2009facilitation}. Moderators enforce rules and ensure orderly interactions, usually with the threat of disciplinary action, though they can also act as community leaders \cite{falk-etal-2024-moderation, seering_self_moderation, Cornell_eRulemaking2017}. Facilitators, on the other hand, guide discussions, promote participation and structured dialogue, particularly in online deliberation and education platforms \cite{Asterhan2010OnlineMO}. Despite these distinctions, the terms are sometimes used interchangeably \cite{cho-etal-2024-language, park_et_al_2012_facilitation, kim_et_al_chatbot}, while it is also common for moderators to use facilitation tactics \cite{Cornell_eRulemaking2017, park_et_al_2012_facilitation, kim_et_al_chatbot, cho-etal-2024-language, proactive_moderation}.

\begin{table*}[!t]
    \centering
    \footnotesize
    \renewcommand{\arraystretch}{1.3}
    \begin{tabular}{lp{10cm}}
        \toprule
        \textbf{Concept} & \textbf{Definition and Characteristics} \\
        \toprule
        \textbf{Discussion} & Broad term encompassing informal and formal exchanges, including online discussions in fora. Can involve elements of debate, dialogue, and deliberation. \\
        \hline
        \textbf{Dialogue} & Collaborative interaction aimed at shared understanding and alignment. Emphasizes cooperation rather than competition. Also refers to dialogue systems in NLP (task-oriented or chatbot conversations). \\
        \hline
        \textbf{Deliberation} & Structured discussion focusing on informed decision-making with reasoned argumentation and diverse perspectives. Less about persuasion, more about collective reasoning. \\
        \hline
        \textbf{Debate} & Adversarial interaction where participants aim to persuade or defend positions rather than achieve mutual understanding. Focused on rhetorical effectiveness. \\
        \hline
        \textbf{Thread-style Discussions} & Online discussions structured in tree/thread formats (e.g., Reddit). Can incorporate elements of all rhetorical styles (debate, dialogue, deliberation). \\
        \hline
        \textbf{Discussion Quality} & Subjective measure influenced by cultural background, engagement, and type of discussion. Defined by socio-dimensional aspects of participant experiences. \\
        \hline
        \textbf{Moderation} & Ensures orderly interactions by enforcing guidelines. Moderators can be volunteers or employees, often associated with disciplinary actions. \\
        \hline
        \textbf{Facilitation} & Encourages equal participation and organizes discussion flow. More common in deliberative and educational contexts, though often used interchangeably with moderation. \\
        \bottomrule
    \end{tabular}
    \caption{Definition of terms used in this survey.}
    \label{tab:terminology}
\end{table*}

\paragraph{Pre-moderation and Post-moderation}

Multiple taxonomies have been proposed to describe the temporal dimension of moderation; that is, when moderator action is applied in relation to when the content is visible to the users \cite{veglis_moderation, proactive_moderation}. These taxonomies are very similar to each other, and usually boil down to the following distinctions:

\begin{itemize}
    \item \textit{Pre-moderation:} The user is dissuaded, or prevented from, posting harmful content. Pre-moderation techniques can include nudges at the writing stage \cite{argyle2023}, reminders about platform rules \cite{proactive_moderation}, or even a moderation queue where posts have to be approved before being visible to others \cite{proactive_moderation}. 
    
    \item \textit{Real-Time:} The moderator is part of the discussion and intervenes like a referee would during a match.
    
    \item \textit{Ex-post:} The moderator is called after a possible incident has been flagged and makes the final call. 
\end{itemize}


\paragraph{Discussion, Deliberation, Dialogue, Debate} 
There is little to no consensus on how to properly define terms such as ``discussion'' and ``dialogue'' \cite{Russman16, Goni2024}.
In this section, we attempt to disambiguate the use of such terms for the purposes of our survey and based on the existing related work. First, our study focuses on \textbf{discussions}, a broader term encompassing various informal and formal exchanges, including online discussions in fora \cite{Russman16}, with which we are mainly concerned. In contrast, \textbf{dialogue} refers to collaborative interactions in which participants work toward a shared understanding and alignment \cite{Redwood18, bawden-2021-understanding,Goni2024}. Studies on dialogue emphasize its cooperative nature, aiming for mutual insight rather than competition \cite{bawden-2021-understanding}. Dialogue can also refer to dialogue systems, a major NLP sub-area,  traditionally including both task-oriented dialogues and casual conversation (Eliza-like)\footnote{\url{http://web.njit.edu/~ronkowit/eliza.html}} ``chatbots'' \cite{ye23, sun-etal-2021-adding}.

A more specific concept is \textbf{deliberation}, which involves structured discussions aimed at informed decision-making, often prioritizing reasoned argumentation and the consideration of diverse perspectives \cite{DEGELING2015114, Lo_McAvoy_2023}. Meanwhile, \textbf{debate} is typically adversarial, where participants focus on persuading others or defending their positions. Unlike dialogue or deliberation, debate centers more on winning or convincing, making it less about collective reasoning and more about rhetorical effectiveness \cite{Lo_McAvoy_2023}. Debates also typically have much stricter (and enforced) rules than other discussions.

For this study, we specifically focus on online written discussions, particularly those occurring in \textbf{thread- or tree-style} formats \cite{seering_self_moderation}. A thread is a collection of messages or posts grouped together in an online forum, discussion board, or messaging platform (such as Reddit). It begins with an initial post (often called the original post, or OP), and subsequent replies are ordered either chronologically or by relevance. Threads usually address a specific topic or question and allow users to engage in discussions about that subject. A thread may grow as users contribute more responses. It must be noted, however, that this type of discussion can contain elements from all the other discussion styles. For example, the adversarial element of the debates, or the argumentative element that can be found both in dialogues and deliberations. 

\paragraph{Discussion Quality} The success of a discussion is often subjective, influenced by a variety of factors such as the cultural background and linguistic proficiency of the participants \cite{zhang-etal-2018-conversations}, as well as their level of engagement \cite{see-etal-2019-makes}. It also depends on the type of the discussion, since
some types of discussions, such as deliberations or debates, may not aim at consensus. Given these complexities, we adopt the definition proposed by \citet{prabhu21}, which views the perceived \textit{discussion quality} as a measurement that attempts to quantify interactions by taking into account multiple socio-dimensional aspects of individual experiences and abilities.

\section{Methodology}


The search and article selection of this survey was conducted using specific keywords in academic search engines (e.g., Google Scholar, Semantic Scholar, Scopus), digital libraries and repositories (e.g., ACL Anthology, ACM Digital Library, IEEE Xplore, JSTOR). We focused on peer-reviewed publications written in English between 2014 and 2024, granting exceptions only for established works predating this period. Additionally, we reviewed other cited papers that appeared highly relevant, provided they were peer-reviewed and cited by more than 20 citations of other researchers, unless the topic was very niche, in which case we judged by its content. The search strategy incorporated keywords and phrases related to \acp{LLM}, discussion facilitation, and discussion evaluation. The list of keywords used is provided
in Table~\ref{tab:keywords}. 
The search was further informed by existing survey articles, such as those by \citet{vecchi-etal-2021-towards} and \citet{wachsmuth-etal-2024-argument}, which served as starting points both for identifying relevant literature and for specifying the vocabulary used in the keyword search.

\section{Discussion Quality Taxonomy}
In this part of the Appendix, we present a table summarizing the discussion evaluation taxonomy (\S\ref{sec:eval_methods}). The dimensions are outlined alongside both pre-\ac{LLM} and \ac{LLM}-based approaches, while also highlighting their respective contributions to facilitation. The dimensions are color-coded for clarity, with orange indicating associated dimensions that could serve as early signs of potential derailment, green marking signs of constructive growth—i.e., conversations going well or worth participating in—and pink denoting interaction dynamics.

\begin{table*}[!t]
\centering
\small
\begin{adjustbox}{width=\textwidth}
\begin{tabular}{|>{\raggedright}p{3.2cm}|>{\raggedright}p{4cm}|>{\raggedright}p{4.2cm}|>{\raggedright\arraybackslash}p{4.2cm}|}
\hline
\cellcolor{blue!25}\textbf{Dimension} & \cellcolor{blue!25}\textbf{Facilitation Use} & \cellcolor{blue!25}\textbf{Pre-LLM Approaches} & \cellcolor{blue!25}\textbf{LLM Approaches} \\
\hline

\rowcolor{taxonomycolor} \multicolumn{4}{|l|}{\textbf{Structure \& Logic}} \\
\cellcolor{peach}\textbf{Argument structure \& \newline analysis} &  \cellcolor{peach}Spot claim-evidence chains; raise early-warning flags; keep debate fact-centred &  \cellcolor{peach}Argument-mining pipelines: claim/premise detection; AQ scoring; graph \& neural models &  \cellcolor{peach}Zero/few-shot AQ labelling; argument-structure parsing; on-the-fly argument-map summaries \\
\hline
\cellcolor{blushpink}\textbf{Coherence \& flow} & \cellcolor{blushpink}Detect topic drift; redirect or bridge gaps & \cellcolor{blushpink}Entity-grid \& sequential coherence models; topic modelling; dialogue state tracking & \cellcolor{blushpink}Prompted coherence scoring; chain-of-thought flow checks; off-topic suggestions \\
\hline
\cellcolor{blushpink}\textbf{Turn-taking} & \cellcolor{blushpink}Monitor balance (entropy/Gini); nudge silent voices; avoid dominance & \cellcolor{blushpink}Turn-entropy / Gini metrics; rule-based alarms & \cellcolor{blushpink}Context-window turn counts; balanced-participation prompts \\
\hline
\cellcolor{peach}\textbf{Linguistic markers} & \cellcolor{peach}Track hedges, 2nd-person spikes, jargon; trigger clarification or civility nudges & \cellcolor{peach} Lexicon features; n-gram-based hedging detectors & \cellcolor{peach}Style-transfer rephrasers; embedding hedge detection; tone-repair suggestions \\
\hline
\cellcolor{blushpink}\textbf{Speech \& dialogue acts} & \cellcolor{blushpink}Identify interruptions, proposals, question types; score deliberative quality & \cellcolor{blushpink}Dialogue-act tagging with ISO/DAMSL labels & \cellcolor{blushpink}Few-shot Dialogue Act tagging; tactic selection based on Dialogue Act patterns \\
\hline
\cellcolor{blushpink}\textbf{Pragmatic comprehension} & \cellcolor{blushpink}Resolve implicatures \& sarcasm; surface hidden misunderstandings & \cellcolor{blushpink}Commonsense reasoning (Knowledge Base + neural); limited coverage & \cellcolor{blushpink}In-context reasoning; auto clarifying questions \\
\hline

\rowcolor{taxonomycolor} \multicolumn{4}{|l|}{\textbf{Social Dynamics}} \\
\cellcolor{peach}\textbf{Politeness} & \cellcolor{peach}Forecast derailment; issue civility nudges or positive reinforcement & \cellcolor{peach}Politeness lexicons; domain-independent classifiers & \cellcolor{peach}Annotation \& polite rewrites; policy-violation explanations \\
\hline
\cellcolor{blushpink}\textbf{Power \& status} & \cellcolor{blushpink}Detect dominance; invite low-status voices; rebalance floor & \cellcolor{blushpink}Style-matching, pronoun analysis; social-role features & \cellcolor{blushpink}Power imbalance estimation; moderator suggestions \\
\hline
\cellcolor{peach}\textbf{Disagreement} & \cellcolor{peach} Distinguish constructive vs destructive dissent; de-escalate & \cellcolor{peach}Graham-hierarchy / stance detection & \cellcolor{peach}Few-shot labelling; automatic reframing prompts \\
\hline

\rowcolor{taxonomycolor} \multicolumn{4}{|l|}{\textbf{Emotion \& Behavior}} \\
\cellcolor{pastelblue}\textbf{Empathy} & \cellcolor{pastelblue}Encourage empathic turns; highlight emotional cues & \cellcolor{pastelblue}Lexicon/coding empathy classifiers; affective features & \cellcolor{pastelblue}Perceived-empathy scoring; supportive paraphrases \\
\hline
\cellcolor{peach}\textbf{Toxicity} & \cellcolor{peach}Flag harmful language; decide moderation step & \cellcolor{peach}BERT/toxicity classifiers; detox lexicons & \cellcolor{peach}Detection + rewrite suggestions; policy chat \\
\hline
\cellcolor{blushpink}\textbf{Sentiment} & \cellcolor{blushpink}Track emotional climate; intervene at negativity spikes & \cellcolor{blushpink}Lexicon \& neural sentiment analysis & \cellcolor{blushpink}Prompt-based labelling; tone-shift detection \\
\hline
\cellcolor{blushpink}\textbf{Controversy} & \cellcolor{blushpink}Sense polarization; invite balancing views & \cellcolor{blushpink}Topic-polarity metrics; ideology models & \cellcolor{blushpink}Ideology tagging; polarity-aware summaries \\
\hline
\cellcolor{pastelblue}\textbf{Constructiveness} & \cellcolor{pastelblue}Stream score; escalate or summarize based on trend & \cellcolor{pastelblue}Feature-based classifiers (linguistic, discourse) & \cellcolor{pastelblue}Constructive-rewrite coaching \\

\rowcolor{taxonomycolor} \multicolumn{4}{|l|}{\textbf{Engagement \& Impact}} \\
\hline
\cellcolor{pastelblue}\textbf{Engagement} & \cellcolor{pastelblue}Detect lulls or dominance; prompt interaction & \cellcolor{pastelblue}Turn/word counts; reply-time gaps & \cellcolor{pastelblue}Auto-recaps; invite quiet users \\
\hline
\cellcolor{pastelblue}\textbf{Persuasion} & \cellcolor{pastelblue}Spotlight evidence-based arguments; dampen manipulation & \cellcolor{pastelblue}Lexical overlap; ethos/pathos/logos; persuasion prediction & \cellcolor{pastelblue}Outcome prediction; neutral framing suggestions \\
\hline
\cellcolor{pastelblue}\textbf{Diversity \& Informativeness} & \cellcolor{pastelblue}Monitor viewpoint spread \& info density & \cellcolor{pastelblue}Topic-diversity indices; IR-based scoring & \cellcolor{pastelblue}Simulate perspectives; propose links \\
\hline

\end{tabular}
\end{adjustbox}
\caption{Summary of discussion quality dimensions and corresponding pre-LLM and LLM-based facilitation strategies.}
\label{tab:discussion_quality_grouped_alt}
\end{table*}

\section {Online Discussion Example with Color-coded Politeness Markers}\label{app:eval_example}

\renewcommand{\arraystretch}{1.3}
\rowcolors{2}{gray!10}{white}
\setlength{\tabcolsep}{10pt}

\begin{table*}[h!]
\centering\small
\begin{tabularx}{\textwidth}{>{\RaggedRight\arraybackslash}p{0.07\textwidth} X}
\rowcolor{gray!30}
\textbf{Turn} & \textbf{Utterance} \\
\hline
0 & \textcolor{green}{Why} \textcolor{blue}{should} \textcolor{orange}{we} help people based on race, and say “\textcolor{orange}{we}’ll help everyone who’s black, because they \textcolor{blue}{could} be poor” instead of just “\textcolor{orange}{we}’ll help everyone who’s poor, in which black people make up a proportionally larger amount”? \\
1 & That study is \textcolor{red}{worse than useless} unless it also distinguishes between “black sounding” names that are associated with wealth and poverty. \\
2 & That wouldn't discount it, that would just add another intersectional axis to investigate.  \&gt;which \textcolor{orange}{I} know without looking that it didn't. \textcolor{green}{How} \textcolor{magenta}{rational}. \\
3 & It's certainly more \textcolor{magenta}{rational} than unquestioningly swallowing everything \textcolor{orange}{I} read, as some people do.  \textcolor{green}{Did} this study of \textcolor{purple}{yours} also test \textcolor{red}{difficult} to pronounce Polish names, or Russian names? Or would that have interfered too much with the foregone conclusion they were attempting to reach? \\
4 & \textcolor{green}{Are you implying that’s what I have done?} You may be the only one making \textcolor{red}{assumptions} here. \\
\end{tabularx}
\caption{Dissucssion example from the Reddit Change My View dataset \cite{chang-danescu-niculescu-mizil-2019-trouble}. Color indicates politeness-related features: \textcolor{blue}{hedging}, \textcolor{orange}{1st person reference}, \textcolor{purple}{2nd person reference}, \textcolor{green}{direct questions}, \textcolor{red}{negative sentiment} and \textcolor{magenta}{positive sentiment}. The annotation was produced with a soon-to-be-released annotation toolkit for discussion evaluation.}\label{app:tab:eval_example}
\end{table*}

Table~\ref{app:tab:eval_example} highlights key politeness-related linguistic features such as hedging, personal references, sentiment, and direct questions. These features are essential in the context of facilitation, where the goal is to guide conversations constructively, maintain safety, and foster mutual understanding. By identifying these elements, the facilitator (human or automatic) can better interpret the tone, intent, and emotional weight of each utterance. For example, detecting hedging or positive sentiment can guide the model to adopt a more collaborative tone, while recognizing negative sentiment or accusatory second-person references may prompt it to de-escalate tension and encourage constructive dialogue. 

\end{document}